\documentclass[11pt]{article}

\usepackage{amsmath,amsbsy,amsfonts,amssymb,amsthm,dsfont,fullpage,units}
\usepackage{xspace}
\usepackage{graphicx}
\usepackage{algorithm,algorithmic,mathtools}
\usepackage{color,cases}
\usepackage{tikz}
\usetikzlibrary{calc,shapes}
\usepackage{subfigure}
\usepackage{multirow}
\usepackage{makecell}
\usepackage{rotating}
\usepackage{bbding}

\usepackage[strict]{changepage}
\usepackage{afterpage}
\usepackage[utf8]{inputenc} %
\usepackage[T1]{fontenc}    %
\usepackage{hyperref}       %
\usepackage{fancyvrb}
\usepackage{xcolor}
\definecolor{LightGray}{gray}{0.98}
\usepackage{url}            %
\usepackage{booktabs}       %
\usepackage{amsfonts}       %
\usepackage{nicefrac}       %
\usepackage{microtype}      %
\usepackage{xcolor}         %
\usepackage{xspace}
\usepackage{caption}
\usepackage{multirow}
\usepackage{tablefootnote}
\usepackage{tikz}
\usepackage{xfrac}
\usepackage{soul}
\usepackage{placeins}
\usetikzlibrary{arrows.meta}
\usetikzlibrary{shapes,arrows,chains,positioning}
\hypersetup{
    colorlinks=true,
    linkcolor=blue,
    urlcolor=blue,
    citecolor=blue,
}

\providecommand{\keywords}[1]
{
  \small	
  \textbf{\textit{Keywords---}} #1
}

\usepackage{tabularx}

\usepackage{hyperref}
\hypersetup{
colorlinks = true,
citecolor = {blue},
citebordercolor = {white}
}
\usepackage{amssymb}
\usepackage{pifont}
\usepackage{units}
\usepackage{amsmath}
\usepackage{amsthm}
\usepackage{graphicx}
\usepackage{esint}
\usepackage{bm}

\def\eg{\emph{e.g.~}}

\usepackage{url}
\usepackage{amsfonts}
\usepackage{nicefrac}
\usepackage{microtype}
\usepackage{xcolor}

\title{Qualitative Failures of Image Generation Models\\ and Their Application in Detecting Deepfakes}

\author{
Ali Borji\\
  \texttt{aliborji@gmail.com} \\
}

\begin{document}

\maketitle

\vspace{-15pt}
\begin{abstract}
The remarkable advancement of image and video generation models has led to the creation of exceptionally realistic content, posing challenges in differentiating between genuine and fabricated instances in numerous scenarios. However, despite this progress, a gap remains between the quality of generated images and those found in the real world. To address this, we have reviewed a vast body of literature from both academic publications and social media to identify qualitative shortcomings in image generation models, which we have classified into five categories. By understanding these failures, we can identify areas where these models need improvement, as well as develop strategies for detecting generated images and deepfakes. The prevalence of deepfakes in today's society is a serious concern, and our findings can help mitigate their negative impact. In order to support research in this field, a collection of instances where models have failed is made available at \href{https://drive.google.com/file/d/1VA1hhlyZ9VqtbGfiIXiqkVfwRvs4p-aJ/view?usp=sharing}{here}.
\end{abstract}


\vspace{-10pt}
\renewcommand{\contentsname}{Table of Content}

{
  \hypersetup{linkcolor=blue}
  \tableofcontents
}

\section{Introduction}




Generated images, also known as synthetic images, are created by machine learning algorithms or other software programs, while real images are captured by cameras or other imaging devices. Generated images are not real-world representations of a scene or object, but rather computer-generated approximations. As such, they lack the authenticity and realism of real images. Deepfakes refer to fabricated media content that has undergone digital alterations to effectively substitute the appearance of one individual with that of another, creating a highly convincing outcome. This paper investigates the indicators that can be utilized for identifying artificially generated images, with a specific focus on detecting deepfakes.

Despite the abundance of anecdotal evidence shared on social media regarding the weaknesses of image generation models, there has yet to be a comprehensive and systematic analysis of these failures. Often, the examples shared by people are selectively chosen to showcase instances in which the models perform well, which may lead to a biased perception of their capabilities, and an overestimation of their effectiveness. While there have been quantitative studies aimed at evaluating and comparing generative models~\cite{borji2019pros,borji2022pros}, such as the use of metrics like FID~\cite{heusel2017gans}, these measures can be difficult to interpret and are usually calculated over large datasets, making them unsuitable for determining the authenticity of individual images. Quantitative measures for detecting deepfakes do exist~\cite{nguyen2022deep}, but they are not as easily accessible to the general public as qualitative measures, which are simpler to carry out.

\begin{figure}[!t]
    \centering
    \includegraphics[width=.9\linewidth]{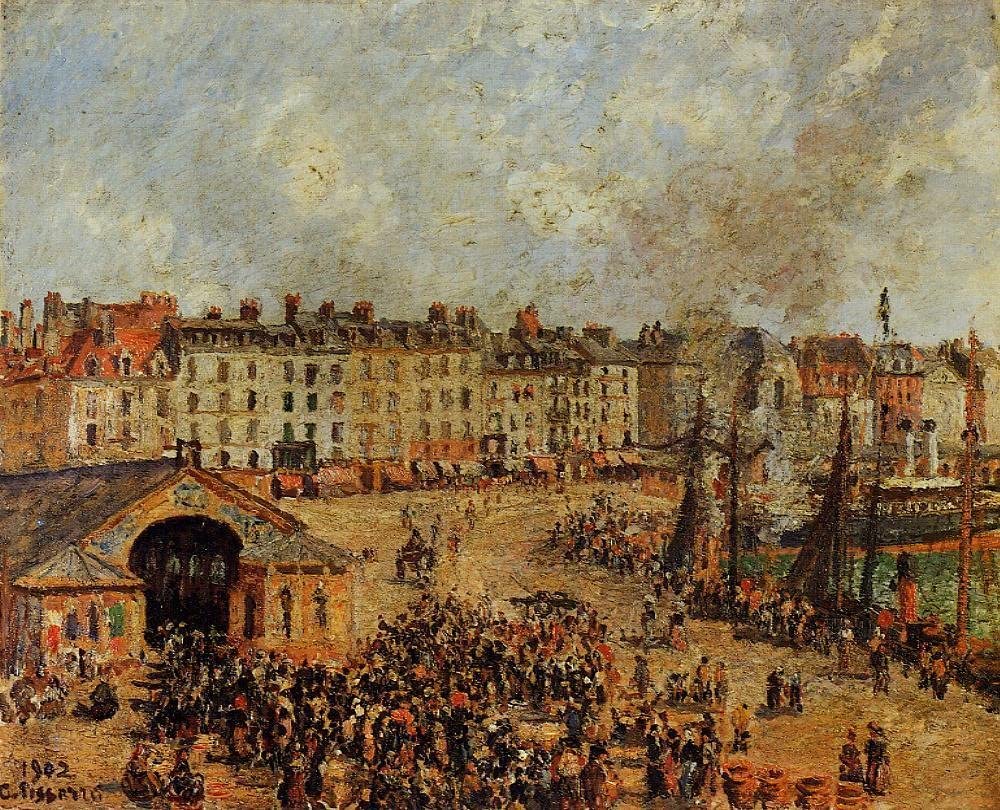} 
    \caption{The Fishmarket, Dieppe, 1902 - Camille Pissarro. When observed more closely, it becomes apparent that the faces in the image lack clarity and numerous details are either incorrect or absent, similar to fake images. Although such images may appear authentic at first glance, scrutinizing them thoroughly is crucial to avoid overlooking errors. It is advisable to conduct a detailed examination of each object within the image by zooming in and analyzing its shape, features, location, and interaction with other objects. This approach allows for a more accurate assessment of the image's authenticity and being free from errors.}
    \label{fig:teaser}
\end{figure}

As the quality of generated images continues to improve, it is crucial to conduct more in-depth and precise analyses. Thus far, people have been amazed by the ability of synthesized images to approximate natural scenes. When Photoshop was introduced, significant efforts were made to identify manipulated images, and a similar approach is needed for generated images today. It would be beneficial to compile a set of indicators and other resources to aid in detecting generated images and deepfakes.

We present a collection of indicators that can be examined in a single image to determine whether it is genuine or generated. Overall, we offer five classes of these indicators including \emph{Human and Animal Body Parts}, \emph{Geometry}, \emph{Physics}, \emph{Semantics and Logic}, as well as \emph{Text, Noise, and Details}, for both portraits and natural landscapes. The advantage of utilizing qualitative cues is that they are easily accessible and can be utilized by anyone, potentially serving as the initial step in detecting deepfakes.

Generated images can appear realistic when viewed from a distance or at high resolutions, making it difficult to discern them from actual photographs. However, at lower resolutions, nearly all generated images lack distinguishable characteristics that set them apart from real photographs. To illustrate, refer to Figure~\ref{fig:teaser}, which depicts a painting by Camille Pissarro featuring intricate details. While the overall image may seem satisfactory, closer inspection reveals several missing details such as distorted facial features.

This study has a dual purpose. Firstly, it aims to explore the differences between generated images and real-world images. Therefore, this research complements studies that propose quantitative approaches for evaluating generative models. Secondly, it aims to examine qualitative methods that can be employed to identify deepfakes and train individuals to become proficient in this task, with the added benefit of systematically organizing this knowledge.

\section{Related work}
\label{related}

\subsection{Quantitative and Qualitative Approaches to Evaluate Generative Models} 
{\bf Quantitative approaches} have emerged as a vital tool to evaluate the performance of generative models. These methods rely on quantitative measures to assess how well a model is able to generate realistic data. One commonly used metric is the Inception Score~\cite{salimans2016improved}, which evaluates the diversity and quality of generated images based on the classification accuracy of a pre-trained classifier. Another popular approach is the Fr\'echet Inception Distance~\cite{heusel2017gans}, which uses feature statistics to compare the distribution of generated data with that of real data. Moreover, other metrics such as precision and recall~\cite{sajjadi2018assessing} can be used to evaluate the quality of generated samples in specific domains such as vision, text and audio. Some studies have proposed methods to assess the visual realism of generated images (\eg~\cite{dragar2023beyond}). These quantitative approaches provide a rigorous and objective way to measure the effectiveness of generative models, helping researchers to improve their models and develop more advanced generative techniques. 

Recently, two metrics have gained popularity, namely the CLIP score and the CLIP directional similarity (\eg~\cite{radford2021learning,ramesh2022hierarchical}). The CLIP score evaluates the coherence of image and caption pairs by measuring their compatibility. A higher CLIP score indicates a greater degree of compatibility, which can also be interpreted as the semantic similarity between the image and the caption. Moreover, studies have shown that the CLIP score has a strong correlation with human judgement. On the other hand, the CLIP directional similarity is used for generating images based on text prompts while being conditioned on an input image. It assesses the consistency between the differences in the two images (in CLIP space) and the differences in their respective captions.

To obtain a thorough analysis of quantitative metrics for evaluating generative models, please refer to the following references~\cite{borji2019pros,borji2022pros,wang2020cnn,zeng2017statistics}.

{\bf Qualitative assessment} of generated images entails a human evaluation. The quality of these images is evaluated on various criteria, such as compositionality, image-text alignment, and spatial relations. 
\href{https://docs.google.com/spreadsheets/d/1y7nAbmR4FREi6npB1u-Bo3GFdwdOPYJc617rBOxIRHY/edit#gid=0}{DrawBench} and \href{https://github.com/google-research/parti}{PartiPrompts} are prompt datasets used for qualitative benchmarking, that are were introduced by Imagen~\cite{saharia2022photorealistic} and Parti~\cite{yu2022scaling}, respectively.
These benchmarks allow for side-by-side human evaluation of different image generation models.

PartiPrompts is a rich set of over 1600 prompts in English. It can be used to measure model capabilities across various categories and challenge aspects such as “Basic”, “Complex”, “Writing \& Symbols”, etc.

\setlength{\tabcolsep}{7pt}
\begin{table}[t]
    \centering
    \resizebox{\textwidth}{!}{
    {\small
    \begin{tabular}{lcc}
    \toprule
    \bfseries{Category} & \bfseries{Description} & \bfseries{Examples}  \\
    \midrule
    \multirow{2}{*}{Colors} & Ability to generate objects & ``A blue colored dog.'' \\
    &  with specified colors. &  ``A black apple and a green backpack.'' \\
    \midrule
    \multirow{2}{*}{Counting} & Ability to generate specified & ``Three cats and one dog sitting on the grass.'' \\
    &  number of objects. & ``Five cars on the street.'' \\
    \midrule
    \multirow{2}{*}{Conflicting} & Ability to generate conflicting & ``A horse riding an astronaut.'' \\
    & interactions b/w objects. & ``A panda making latte art.'' \\
    \midrule
    \multirow{2}{*}{DALL-E \cite{ramesh2021zero}} & Subset of challenging prompts & ``A triangular purple flower pot.'' \\
    & from \cite{ramesh2021zero}. & ``A cross-section view of a brain.'' \\
    \midrule
    \multirow{2}{*}{Description} & Ability to understand complex and long & ``A small vessel propelled on water by oars, sails, or an engine.'' \\
    & text prompts describing objects. & ``A mechanical or electrical device for measuring time.'' \\
    \midrule
    \multirow{2}{*}{Marcus et al. \cite{marcus2022very}} & Set of challenging prompts & ``A pear cut into seven pieces arranged in a ring.'' \\
    & from \cite{marcus2022very}. & ``Paying for a quarter-sized pizza with a pizza-sized quarter.'' \\
    \midrule
    \multirow{2}{*}{Misspellings} & Ability to understand & ``Rbefraigerator.'' \\
    &  misspelled prompts. & ``Tcennis rpacket.'' \\
    \midrule
    \multirow{2}{*}{Positional} & Ability to generate objects with & ``A car on the left of a bus.'' \\
    &  specified spatial positioning. & ``A stop sign on the right of a refrigerator.'' \\
    \midrule
    \multirow{2}{*}{Rare Words} & \multirow{2}{*}{Ability to understand rare words\footnote{https://www.merriam-webster.com/topics/obscure-words}.} & ``Artophagous.'' \\
    & & ``Octothorpe.'' \\
    \midrule
    \multirow{2}{*}{Reddit} & Set of challenging prompts from & ``A yellow and black bus cruising through the rainforest.'' \\
    & DALLE-2 Reddit\footnote{https://www.reddit.com/r/dalle2/}.& ``A medieval painting of the wifi not working.'' \\
    \midrule
    \multirow{2}{*}{Text} & \multirow{2}{*}{Ability to generate quoted text.} & ``A storefront with 'Deep Learning' written on it.'' \\
    & & ``A sign that says 'Text to Image'.'' \\
    \bottomrule
    \end{tabular}
    }
    }
    \vspace*{0.3cm}
    \caption{Description and examples of the 11 categories in DrawBench, compiled from~\cite{saharia2022photorealistic}.}
    \label{tab:drawbench-categories}
\end{table}

DrawBench is comprised of a collection of 200 prompts that are divided into 11 categories (Table~\ref{tab:drawbench-categories}), which aim to assess various capabilities of models. These prompts test a model's ability to accurately render different attributes, such as colors, object counts, spatial relationships, text in the scene, and unusual object interactions. Additionally, the categories include complex prompts that incorporate lengthy, intricate textual descriptions, as well as uncommon words and misspelled prompts. DrawBench was used to directly compare different models, where human evaluators were presented with two sets of images, each consisting of eight samples, one from Model A and the other from Model B. Evaluators were then asked to compare Model A and Model B based on sample fidelity and image-text alignment.

Large-scale datasets have also been used in studies that focus on the qualitative evaluation of generated images (\eg~\cite{assogba2023large}).

The assessment of models through qualitative methods can be susceptible to errors, potentially leading to an incorrect decision. Conversely, quantitative metrics may not always align with image quality. Therefore, the use of both qualitative and quantitative evaluations is typically recommended to obtain a more robust indication when selecting one model over another.

\subsection{Deepfake Detection Methods}

Detection of deepfakes has become an essential area of research due to the increasing sophistication of deep learning algorithms that can generate highly realistic fake images, videos, and audio. As a result, numerous deepfake detection methods have been proposed in recent years, ranging from traditional image and video forensic techniques to advanced deep learning-based approaches. These methods can be broadly categorized into two groups: static and dynamic analysis.

Static analysis methods use handcrafted features to distinguish between real and fake images. Examples of static analysis methods include reverse image search, which compares the content of an image to a large database of known images (\eg~\cite{chen2005content}), and error level analysis, which detects inconsistencies in the compression levels of an image~\cite{kee2011digital}. Another method is the use of noise patterns and artifacts, which are common in images and videos captured by digital cameras and can be used to identify forgeries. For instance, the sensor pattern noise in images captured by digital cameras can be used to authenticate images and detect tampering attempts~\cite{lukas2006digital}. In addition, traditional forensic techniques such as shadow analysis, lighting analysis, and perspective analysis can also be used to identify inconsistencies in the shadows, lighting, and perspectives of images.

On the other hand, dynamic analysis methods rely on deep neural networks to analyze the temporal features of video and audio data to detect deepfakes. These methods aim to exploit the fact that deepfakes lack the natural temporal variations and correlations that are present in real videos and audios. For instance, the use of convolutional neural networks (CNNs) has been proposed to detect deepfakes by analyzing the spatial features of images and videos (\eg~\cite{afchar2018mesonet,mo2018fake,nataraj2019detecting,cozzolino2018forensictransfer}). Similarly, recurrent neural networks (RNNs) have been proposed to analyze the temporal features of video and audio data to detect deepfakes~\cite{guera2018deepfake}. Moreover, Generative Adversarial Networks (GANs)~\cite{goodfellow2020generative} have been used to generate fake images and videos, but can also be used to detect them by identifying inconsistencies in the generator's output~\cite{li2018exposing}.

Overall, deepfake detection is a challenging problem due to the rapid evolution of deep learning algorithms that can generate more realistic fake content~\cite{chesney2019deep}. Thus, a combination of static and dynamic analysis approaches is necessary to achieve effective detection of deepfakes. Additionally, extensive evaluation and comparison of deepfake detection methods are essential to identify their effectiveness and limitations and to guide future research in this area. To read more about this subject, you may want to consult~\cite{fridrich2009digital,redi2011digital,nguyen2022deep,verdoliva2020media} which offer comprehensive reviews on the topic.

\section{Qualitative Failures of Image Generation Models}
We compiled a list of qualitative failures by examining images from various sources including social media websites such as Twitter, LinkedIn, Discord, and Reddit\footnote{A few of the images used in this work were obtained with the consent of a Reddit user named \href{https://www.reddit.com/user/Kronzky}{Kronzky}.}, as well as images from the DiffusionDB dataset~\cite{wang2022diffusiondb}\footnote{This dataset includes prompts that were used to generate images.}. These images have been generated by notable generative models such as \href{https://openai.com/product/dall-e-2}{DALL-E 2}, \href{https://www.midjourney.com/}{Midjourney}, \href{https://stability.ai/}{StableDiffusion}, and \href{https://www.bing.com/images/create/}{Bing Image Creator}. Additionally, we analyzed images from websites such as \href{https://thisxdoesnotexist.com}{thisxdoesnotexist.com}, \href{https://www.whichfaceisreal.com/}{whichfaceisreal.com}, the \href{https://stock.adobe.com/}{Adobe Stock library}, and \href{https://openart.ai/}{openart.ai}. We made sure that the text prompts used to generate images were not intentionally seeking peculiar images. Finally, we manually reviewed the images and filtered out the ones without problems.

\subsection{Human and Animal Body Parts}

\begin{figure}[t]
    \centering
    \includegraphics[width=.9\linewidth]{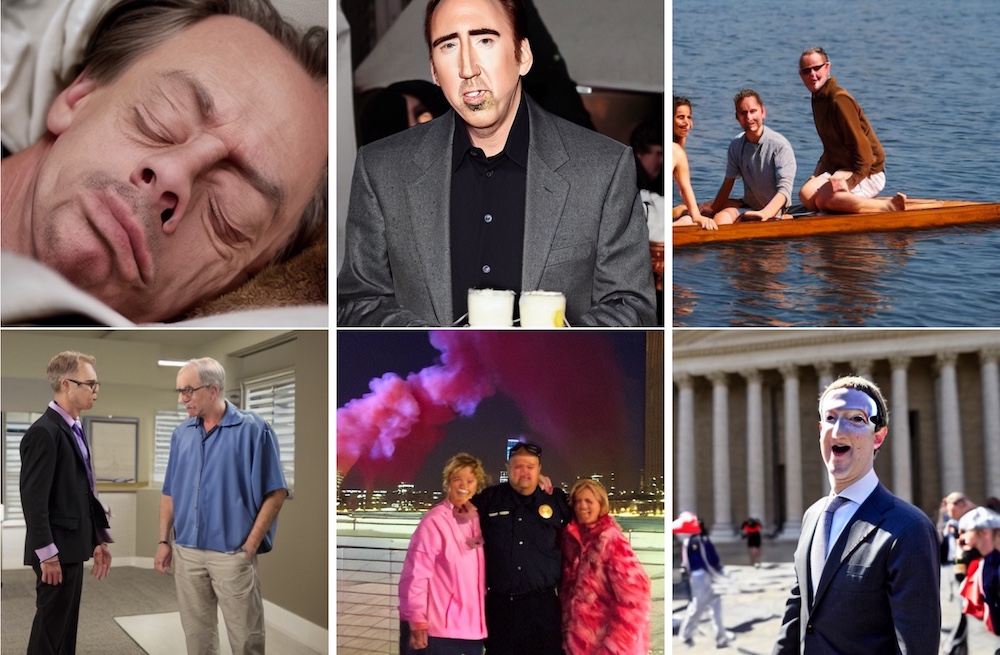}
    \caption{Examples of poorly generated faces.}
    \label{fig:face}
\end{figure}

\noindent {\bf Faces.}
Since the initial triumphs of GANs, the generation of fake faces has been the most extensively scrutinized category for deep generative models~\cite{borji2022generated}. Faces are comparatively simpler to generate than complex scenes because they are easier to calibrate. In the past, the first generated faces were effortlessly recognizable by humans. However, with the advancement of technology such as StyleGAN~\cite{karras2020analyzing}, the latest examples of generated faces are more challenging to distinguish. Figure~\ref{fig:face} illustrates a few faces that were generated with issues. You can evaluate your ability to distinguish between real and computer-generated faces by taking a quiz at \href{https://www.whichfaceisreal.com/}{whichfaceisreal.com}.

\begin{figure}[t]
    \centering
    \includegraphics[width=1\linewidth]{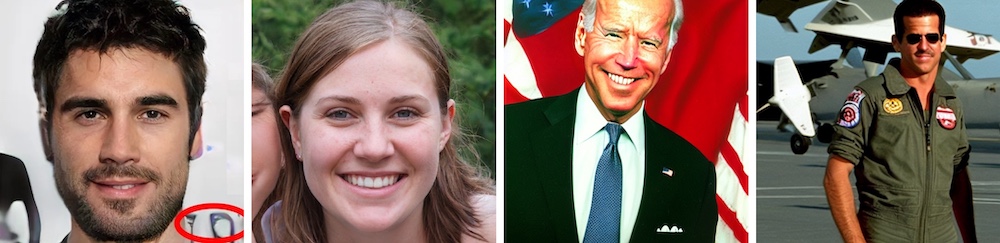}
    \caption{Fake images can be exposed through background cues.}
    \label{fig:background}
\end{figure}

\noindent {\bf Image Background.}
When creating generated images and deepfakes, issues with the background of the images may arise, particularly in cases where the face is in focus while the surrounding clues are incorrect. The neural network used to generate the images focuses mainly on the face and may not pay as much attention to the surrounding details. This can lead to strange companions or chaotic forms in the background. Additionally, the objects or people next to the primary person in the image may appear unnatural or ``mutant". Figure~\ref{fig:background} displays several instances of failures as examples.

\begin{figure}[t]
    \centering
    \includegraphics[width=1\linewidth]{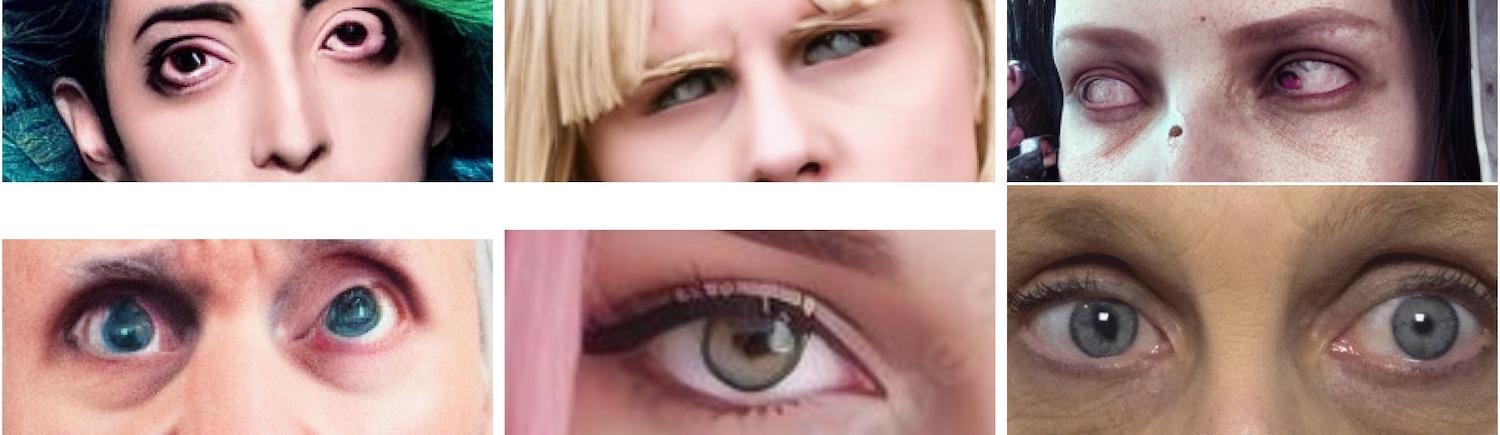}
    \caption{Here are some instances of eyes that were generated poorly. The eye in the bottom right corner is an actual photograph of a patient who has an irregularly shaped pupil. You can refer to \href{https://n.neurology.org/content/91/15/715}{this link} for more details. This case represents a unique manifestation of a condition known as ``cat's eye Adie-like pupil," which is considered a warning sign for ICE syndrome.}
    \label{fig:eye}
\end{figure}

\noindent {\bf Eyes and Gaze.}
Deep generative models have largely overcome issues with early fake images such as cross-eyed, uncentered or different sized pupils, different colored irises, and non-round pupils, as shown in examples in Figure~\ref{fig:eye}. Early GANs used to produce pupils that were not circular or elliptical like those found in real human eyes, which can be a clue that an image is fake. Reflections in the eyes can also be used to identify fake images. Other clues include irregularities in pupil shape, although this is not always indicative of a fake image since some diseases can cause such irregularities. See the example shown in the bottom-right panel in Figure~\ref{fig:eye}.

\begin{figure}[t]
    \centering
    \includegraphics[width=.9\linewidth]{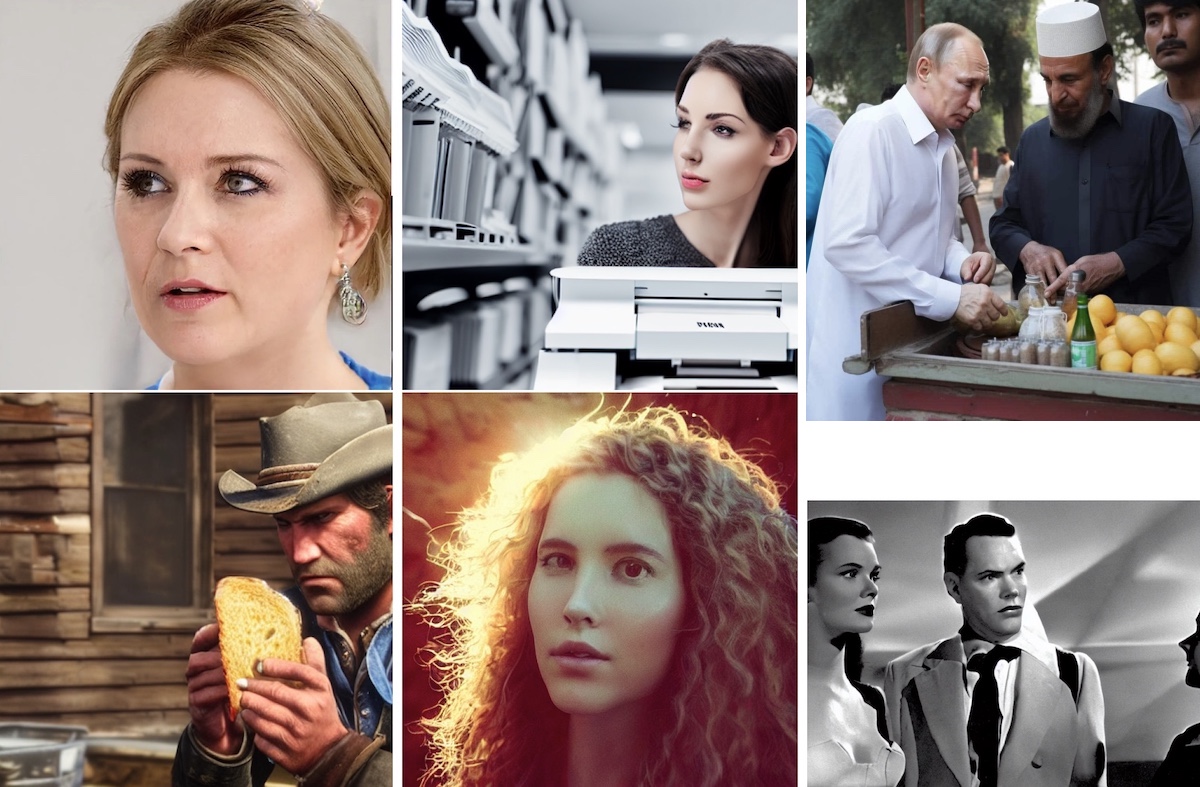}
    \caption{Here are some examples of images where the gaze direction is problematic. In these images, one eye appears to be looking in a different direction compared to the other, similar to a medical condition called Strabismus in the real world. You can check out \url{https://en.wikipedia.org/wiki/Strabismus} for additional information on this topic.}
    \label{fig:gaze}
\end{figure}

Unnatural gaze direction or unrealistic eye movements may be observed in deepfakes, which can indicate that a machine learning algorithm generated or manipulated the image. Please see Figure~\ref{fig:gaze}.

\begin{figure}[t]
    \centering
    \includegraphics[width=.8\linewidth]{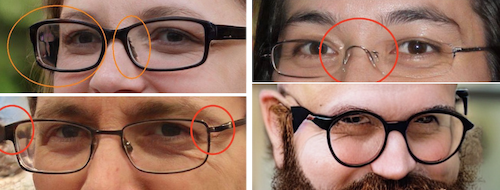}
    \caption{Some samples of generated eyeglasses with poor quality.}
    \label{fig:glasses}
\end{figure}

\noindent {\bf Eyeglasses.}
Algorithms can struggle to create realistic eyeglasses, with frame structures often differing between the left and right sides, or with one side having an ornament and the other not. Sometimes the frame can appear crooked or jagged. The glasses may partially disappear or blend with the head, and they can be asymmetrical. The view through the lens may also be heavily distorted or illogical, and nose pads may be missing or distorted. Please see Figure~\ref{fig:glasses} for some examples.

\begin{figure}[t]
    \centering
    \includegraphics[width=1\linewidth]{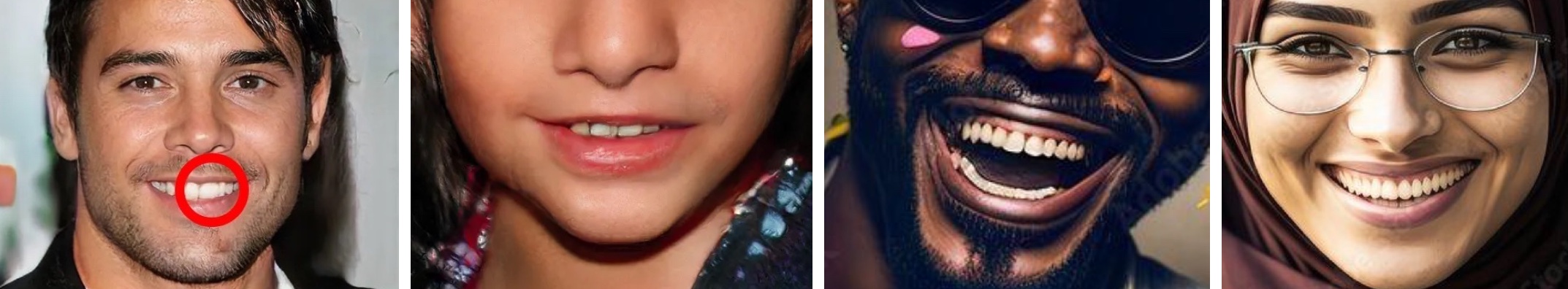}
    \caption{Examples of poorly generated teeth.}
    \label{fig:teeth}
\end{figure}

\noindent {\bf Teeth.}
Rendering teeth is a difficult task for AI, which often results in odd or asymmetric teeth. When someone's teeth appear unusual or crooked, there's a good chance that the image was generated by AI. Semi-regular repeating details like teeth are difficult for models to generate, causing misaligned or distorted teeth. This problem has also been observed in other domains, such as texture synthesis with bricks. Occasionally, an image may display an excessive number of teeth or teeth with abnormal shapes and colors, and in some instances, there may be an insufficient number of incisors. Please see Figure~\ref{fig:teeth} for some examples.

\begin{figure}[!htbp]
    \centering
    \includegraphics[width=.6\linewidth]{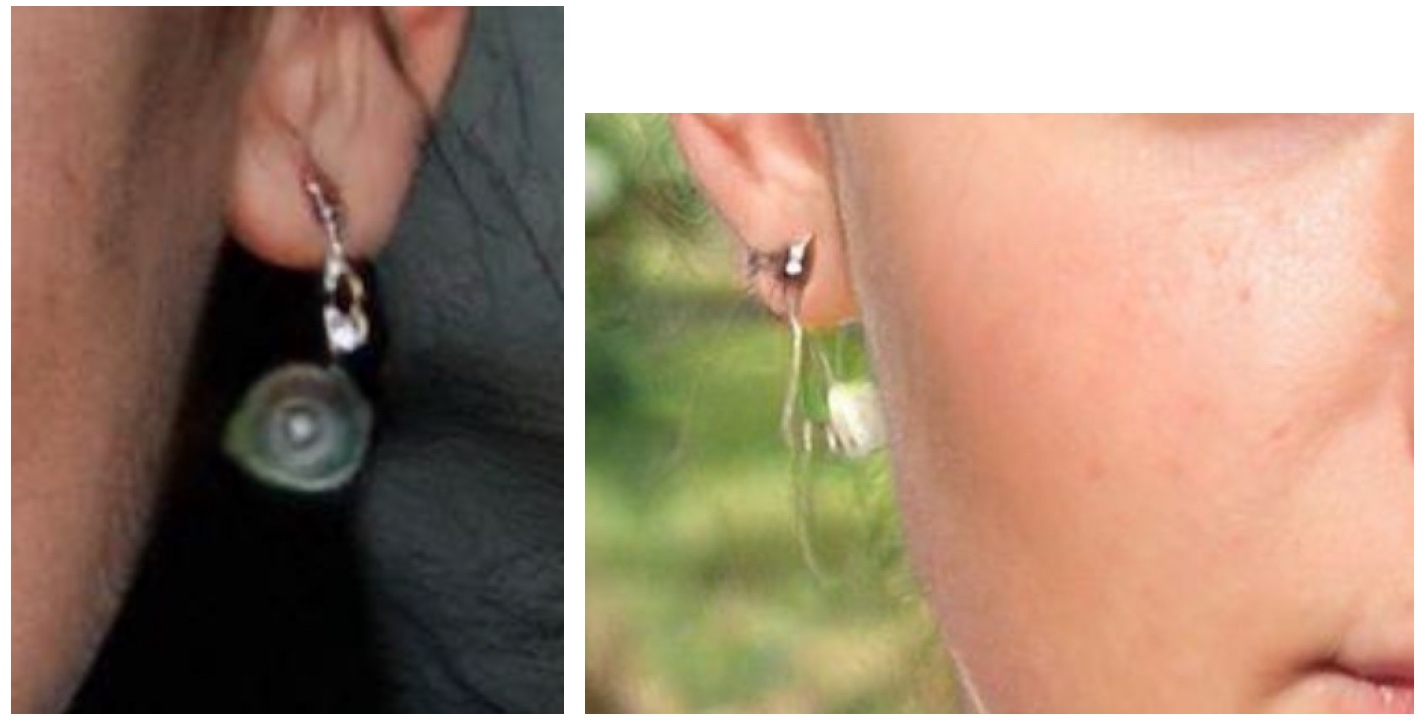}
    \caption{Clues that can reveal fake ears, here through earrings.}
    \label{fig:ear}
\end{figure}

\noindent {\bf Ear and Earrings.}
Ears in AI-generated images may exhibit discrepancies such as differences in size, one ear appearing higher or bigger than the other, or missing or partially missing earrings. Additionally, earrings may be randomly shaped or not match visually. If earrings are asymmetrical or have different features such as one having an attached earlobe while the other doesn't or one being longer than the other, it's likely that the image has been generated by AI. Examples of poorly generated ears and earrings are shown in Figure~\ref{fig:ear}. 

\begin{figure}[!t]
    \centering
    \includegraphics[width=.8\linewidth]{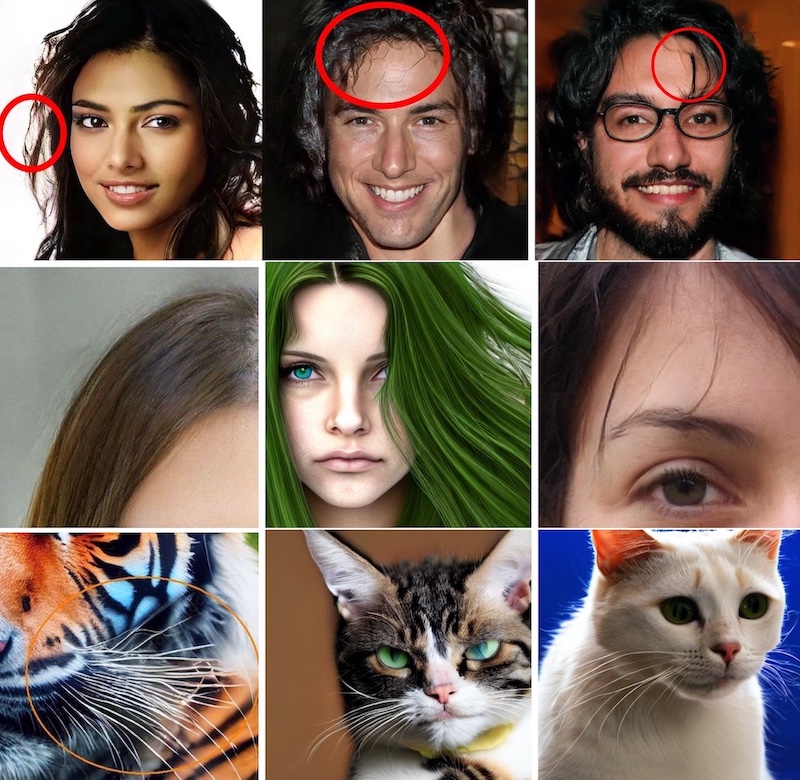}
    \caption{Examples of poorly generated hair.}
    \label{fig:hair}
\end{figure}

\noindent {\bf Hair and Whiskers.}
The style of hair can differ greatly, which also means there is a lot of intricate detail to capture. This makes it one of the most challenging aspects for a model to render accurately. The generated images may contain stray strands of hair in unusual places, or the hair may appear too straight or streaked. Occasionally, the image may resemble acrylic smudges from a palette knife or brush. Another issue may be a strange glow or halo around the hair. In some cases, the model may bunch hair in clumps or create random wisps around the shoulders, while also including thick stray hairs on the forehead. Please see Figure~\ref{fig:hair}.

\begin{figure}[!htbp]
    \centering
    \includegraphics[width=1\linewidth]{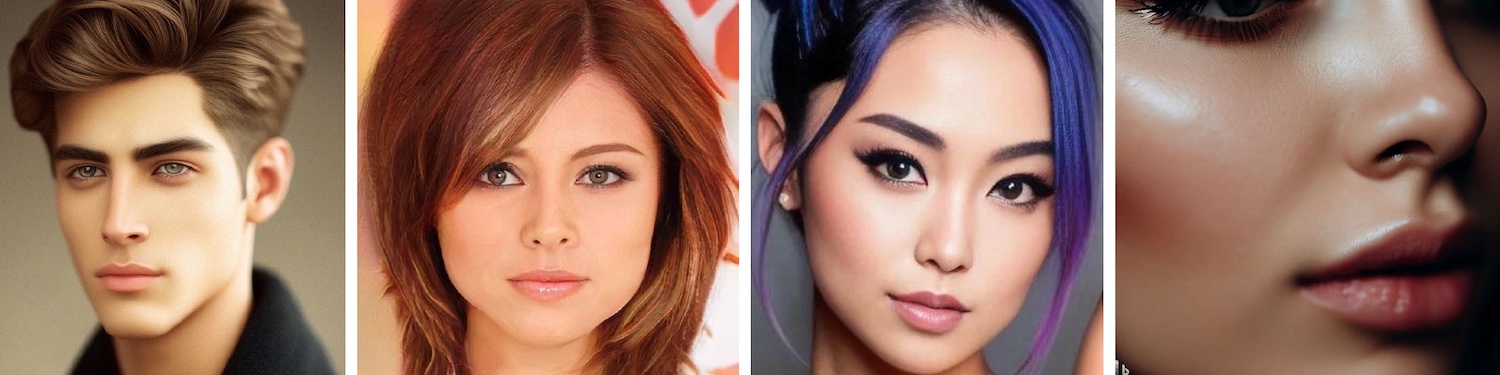}
    \caption{Examples of poorly generated skin, absolutely perfect skin with no pores.}
    \label{fig:skin}
\end{figure}

\noindent {\bf Skin.} 
Deepfakes can be deficient in delicate details and subtleties found in genuine images, like skin texture, pores, or fine lines on someone's face. The skin tone in deepfakes may appear unnatural or inconsistent, such as a person's face appearing too pale or too red. Additionally, deepfakes may lack the presence of noise or grain which exists in real images, giving a sense of texture and realism. Without the presence of noise or grain, deepfake images may seem excessively clean or artificial.
Some example failures are shown in Figure~\ref{fig:skin}.

\begin{figure}[!htbp]
    \centering
    \includegraphics[width=1\linewidth]{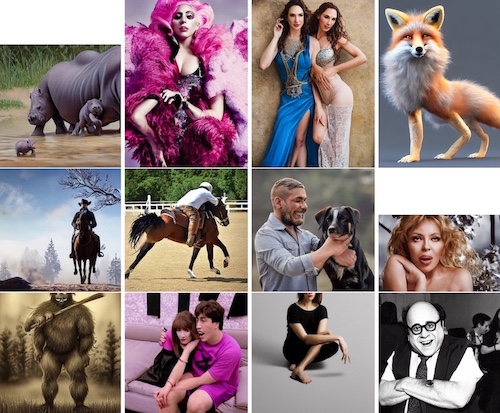}
    \caption{Examples of images with poorly generated limbs and distorted body.}
    \label{fig:limb}
\end{figure}

\begin{figure}[!htbp]
    \centering
    \includegraphics[width=1\linewidth]{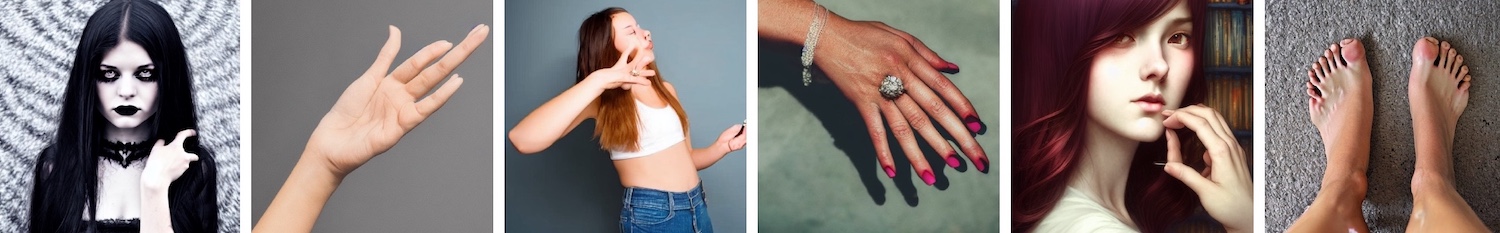}
    \caption{Issues with AI-generated fingers.}
    \label{fig:finger}
\end{figure}

\noindent {\bf Limbs, Hands, and Fingers.}
The models used for generating deepfakes often fall short when it comes to accurately depicting the intricate details of human extremities. For instance, hands may randomly duplicate, fingers can merge together or there may be too many or too few of them, and third legs may unexpectedly appear while existing limbs may disappear without a trace. Furthermore, limbs may be positioned in unrealistic or impossible poses, or there may be an excess number of them. As a result, deepfakes may exhibit unnatural body language, such as unrealistic gestures or postures that are out of place. In certain instances, models are unable to accurately depict the interaction between objects and body parts (\eg brushing).
See Figs~\ref{fig:limb} and~\ref{fig:finger}.

\begin{figure}[!t]
    \centering
    \includegraphics[width=.8\linewidth]{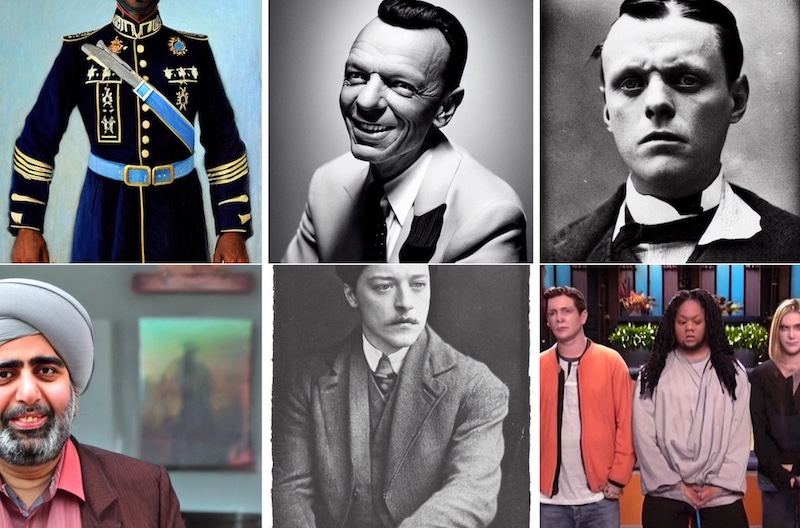}
    \caption{Generating realistic clothing is a challenge for generative models.}
    \label{fig:clothing}
\end{figure}

\noindent {\bf Clothing.} 
Generative models may produce distorted clothing with various issues, such as asymmetrical, peculiar, or illogical textures or components such as zippers or collars merging with the skin, and textures abruptly changing or ending. Please refe to Figure~\ref{fig:clothing} for some of such failures. 

\subsection{Geometry}
Generated images may exhibit anomalous or atypical image geometry, with objects appearing to be of an unusual shape or size, in comparison to their expected proportions.

\begin{figure}[!t]
    \centering
    \includegraphics[width=.8\linewidth]{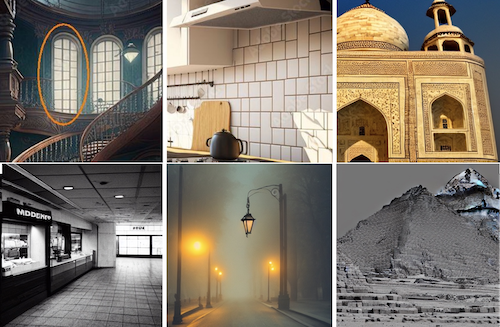}
    \caption{Examples of lines, edges, and surfaces that are generated poorly by AI.}
    \label{fig:edges}
\end{figure}

\noindent {\bf Straight Lines and Edges.}
AI-generated images may lack the straight lines, seams, and connections found in real-world objects, resulting in wavy, misaligned, and jumpy renderings (\eg in tiles). Generated images can also exhibit inconsistent or unnatural image edges, which refer to the boundaries between different parts of the image. Further, surfaces, which are typically straight, may look somewhat uneven in generated images. Some samples failures are shown in Figure~\ref{fig:edges}. 

\begin{figure}[!htbp]
    \centering
    \includegraphics[width=.9\linewidth]{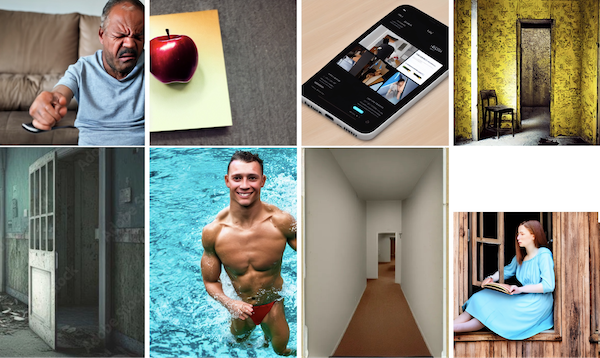}
    \caption{Examples of generated images that exhibit issues with perspective.}
    \label{fig:perspective}
\end{figure}

\noindent {\bf Perspective.}
Models lack the ability to understand the 3D world, which results in physically impossible situations when objects cross different planes in a scene. These errors are difficult to detect as our brain often auto-corrects them, requiring a conscious investigation of each angle of the object to identify inconsistencies. Generated images can display an unnatural or distorted perspective, where a person's body appears stretched or compressed unrealistically. They may also have inconsistent or unrealistic camera angles, where a person's face appears to be viewed from an impossible angle or perspective. Some example failures are shown in Figure~\ref{fig:perspective}.

\begin{figure}[!htbp]
    \centering
    \includegraphics[width=.9\linewidth]{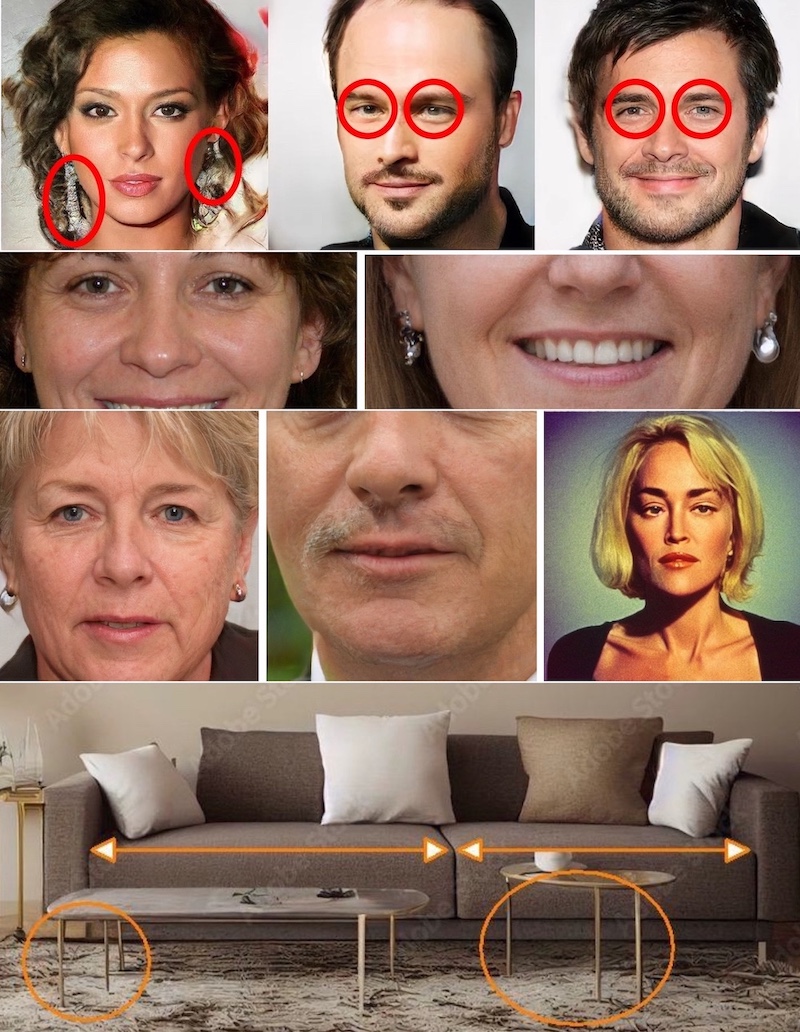}
    \caption{Examples of generated images that display inconsistent symmetry.}
    \label{fig:symmetry1}
\end{figure}

\begin{figure}[!t]
    \centering
    \includegraphics[width=1\linewidth]{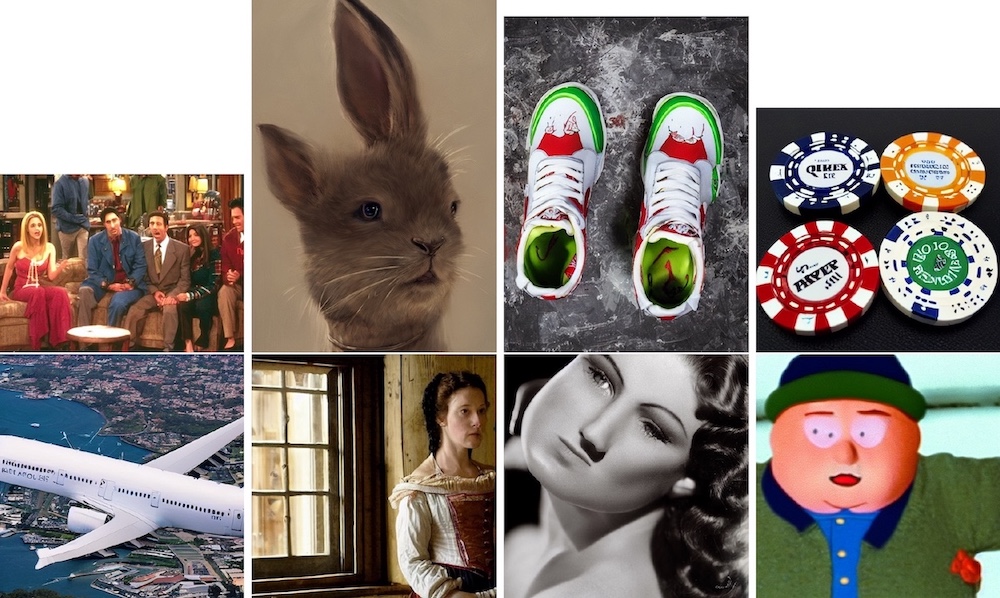}
    \caption{Additional examples of generated images that exhibit inconsistent symmetry.}
    \label{fig:symmetry2}
\end{figure}

\noindent {\bf Symmetry.}
 Due to difficulty managing long-distance dependencies in images, symmetry (reflection, radial, translation, etc) can be challenging for models. For instance, in generated images, eyes may appear heterochromatic and crosseyed, unlike in real life where they tend to point in the same direction and have the same color. Additionally, asymmetry may appear in facial hair, eyeglasses, and the types of collar or fabric used on the left and right sides of clothing. 
Models may face challenges in maintaining symmetry not only in faces but also in other objects and scenes. For instance, two shoes in a pair or wings in an airplane might not be exactly the same. This is a type of reasoning glitch where the model cannot understand that certain elements should be symmetrical. Some example failures are shown in Figures~\ref{fig:symmetry1} and~\ref{fig:symmetry2}.

\begin{figure}[!t]
    \centering
    \includegraphics[width=1\linewidth]{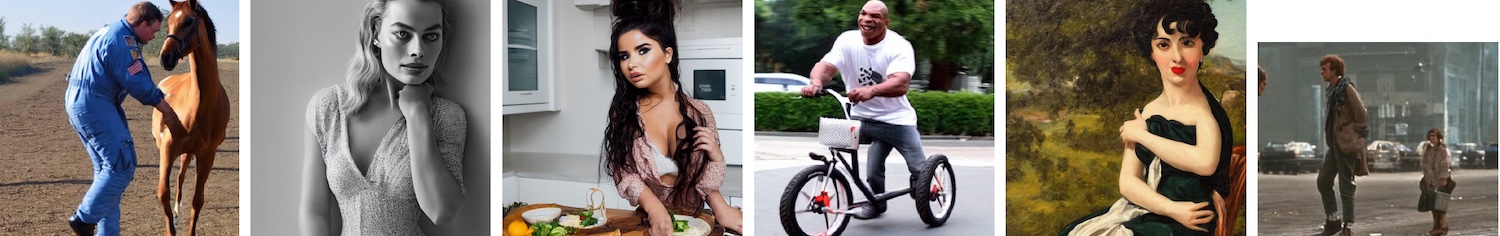}
    \caption{Examples of images where there is a violation of relative size.}
    \label{fig:size}
\end{figure}

\noindent {\bf Relative Size.}
Relative size is a visual perceptual cue that helps us understand the size of objects in relation to one another. It is a powerful cue because it allows us to estimate the size of objects even when we do not have any absolute size reference in the scene.
Models, however, fall short in synthesizing objects with objects with sizes proportional to their size in the real world. Some example failures are shown in Figure~\ref{fig:size}.

\noindent {\bf Other Geometry.} 
Generated images exhibit various geometrical anomalies that may reveal their artificiality. For instance, their depth cues can be inconsistent or unnatural, causing the foreground or background to seem blurry or devoid of detail. Moreover, they often lack parallax, which is the apparent displacement of objects when viewed from different perspectives, resulting in a flat or two-dimensional appearance. Additionally, incorrect or inconsistent motion blur may suggest that certain parts of the image have been manipulated. The absence of occlusion, i.e., the overlapping of objects in the scene, is another telltale sign of generated images, as it can make the image look flat or unrealistic. Lastly, generated images may display improper image alignment, with objects seeming misaligned or out of place.

\subsection{Physics}
Generated images that violate physics rules exhibit various cues that can give them away as unrealistic or physically impossible. These cues include objects appearing to float in mid-air without support, shadows that are inconsistent with the light source, reflections or refractions that break the laws of optics, objects passing through each other without interaction, and incorrect physics-based simulations such as fluids or cloth that behave in impossible ways. By identifying these cues, it is possible to identify and distinguish realistic images from those that violate the rules of physics.

\begin{figure}[!t]
    \centering
    \includegraphics[width=1\linewidth]{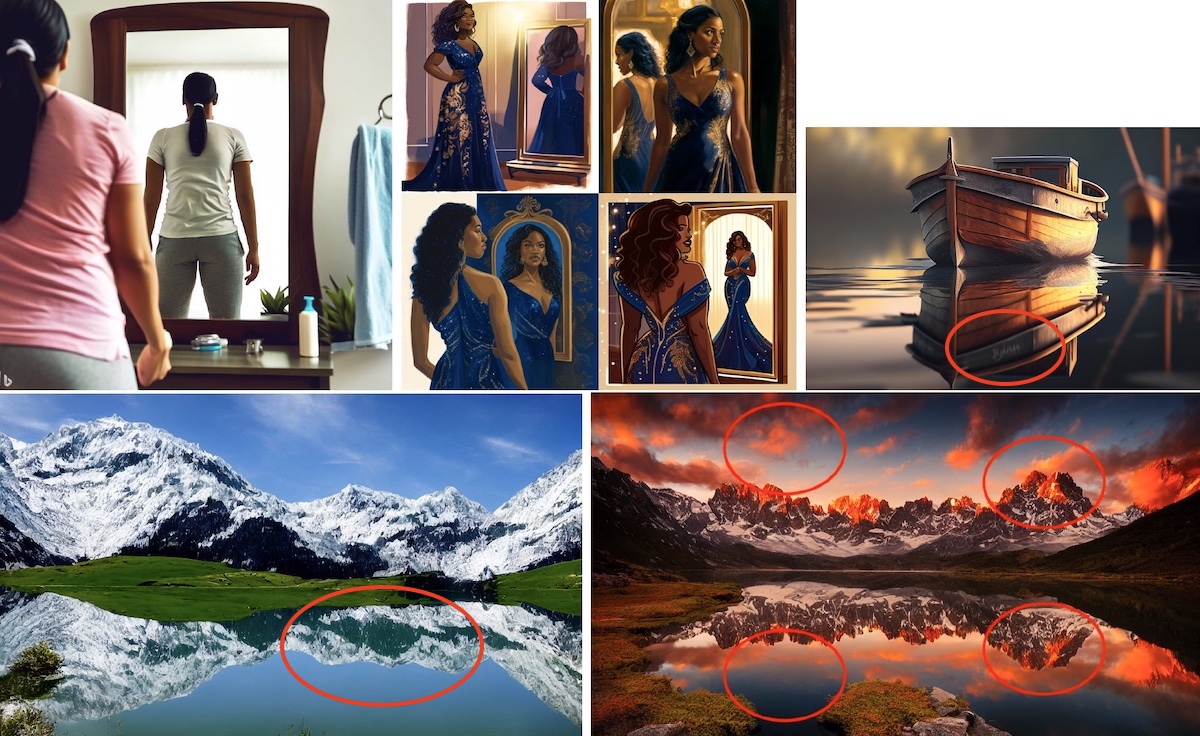}
    \caption{Generated images with inconsistent reflections.}    
    \label{fig:reflection}
\end{figure}

\noindent {\bf Reflection.} 
An effective technique for detecting generated images is to examine the lighting and how it interacts with the elements within the image, and how it causes reflections and shadows. Generated images can exhibit artificial reflections that are inconsistent with the natural lighting and environment, such as those in glasses, mirrors, or pupils. The root cause of this issue is that deep generative models lack a proper understanding of reflections. While these models may recognize that an image contains a reflection and typically involves two people (one facing the camera and the other with their back turned), they do not comprehend that the two individuals are, in fact, the same person. Generated images may display other lighting effects that do not match real-world environments, such as lens flares, lens distortion, chromatic aberration and unnatural specular highlights. These effects are frequently observed in genuine photographs due to the physical properties of camera lenses and the way light is refracted through them. Some example failures are shown in Figure~\ref{fig:reflection}.

\begin{figure}[!t]
    \centering
    \includegraphics[width=1\linewidth]{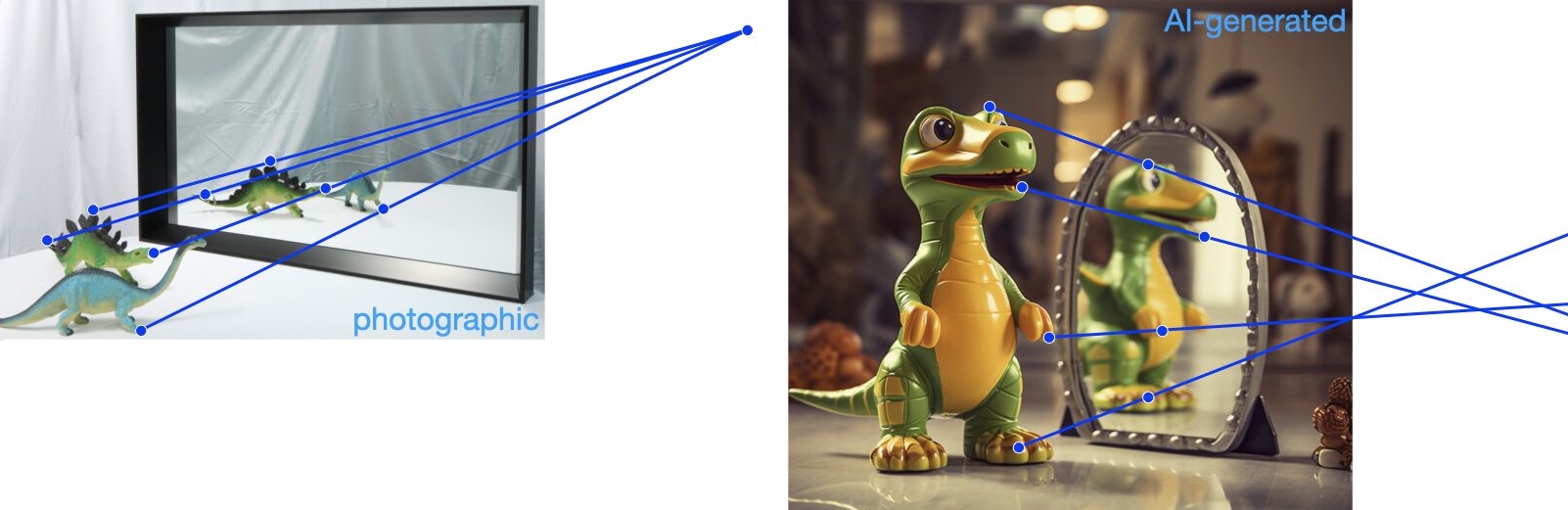}
    \caption{Consistent and inconsistent reflections in real (left) vs. generated images.}
    \label{fig:geometryreflection}
\end{figure}

AI-generated images sometimes exhibit inconsistencies in the geometry of reflections~\cite{farid2022perspective}. Shown in the left panel of Fig.~\ref{fig:geometryreflection} is a photographic image in which lines that connect points on the toy dinosaur in the scene and their reflections in the mirror all converge to a single point (a vanishing point). Reflections in AI-generated images (the right panel in Fig.~\ref{fig:geometryreflection}), however, exhibit physical inconsistencies as can be seen by the lack of a consistent vanishing point.

\begin{figure}[!t]
    \centering
    \includegraphics[width=.9\linewidth]{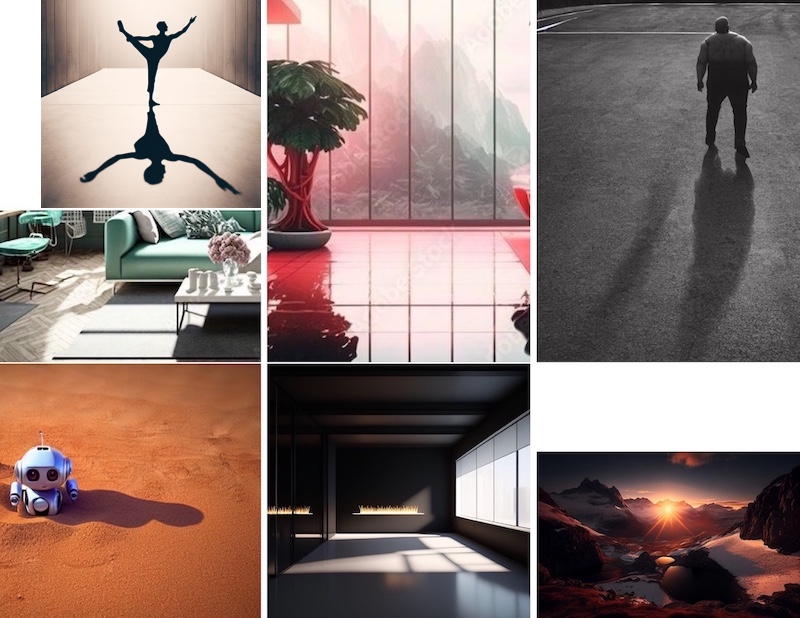}
    \caption{Generated images with inconsistent shadows.}
    \label{fig:shadow}
\end{figure}

\noindent {\bf Shadow.}
Generated images might not include shadows, which are typically found in real images and contribute to the impression of depth and authenticity.
It is important to observe objects without shadows and those with highlights that appear to originate from a different direction than the rest of the image. Additionally, if the photo was taken outdoors in natural light during the afternoon, the setting sun will produce longer shadows than it would at midday, which can be easily identified by scrutinizing the shadow's length. However, this method may not be as precise in artificial lighting conditions. Finally, if there are multiple objects or people within the scene, their shadows should be consistent with each other. Some generated images with inconsistent shadows are shown in~\ref{fig:shadow}.

\begin{figure}[!t]
    \centering
    \includegraphics[width=.9\linewidth]{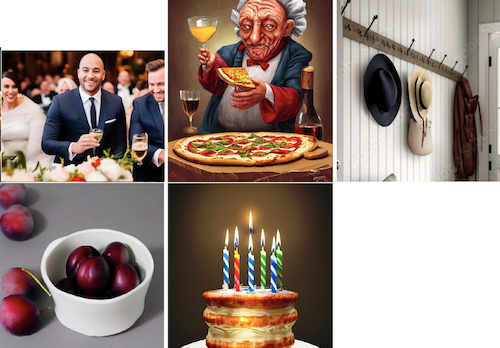}
    \caption{Generated images where some objects lack visible physical support. Some objects are suspended in mid-air without any explanation or justification. This lack of physical support could result from a failure to properly simulate or model the forces acting on the objects in the scene.}
    \label{fig:nosupport}
\end{figure}

\noindent {\bf Objects without Support.}
When an object or material appears to be floating in mid-air without any visible means of support, it gives the impression that the object is defying gravity or the laws of physics. In reality, all objects are subject to the force of gravity unless they are held up by some other force. When an object appears to be floating, it could be a result of an incorrect rendering or an error in the physics simulation that fails to account for the gravitational force. This type of inconsistency can cause a generated image to look unrealistic or implausible. Some example failures are shown Figure~\ref{fig:nosupport}.

\subsection{Semantics and Logic}

Images produced by generative models may lack the semantic meaning or contextual relationships present in authentic images. These models tend to focus on the nouns in a given prompt and construct a plausible scene based on them, potentially failing to capture the true relationships between objects. It is crucial to bear in mind that AI lacks an inherent understanding of the world and can only process information in terms of shapes and colors. Complex concepts, such as logical connections and three-dimensional space, are beyond its grasp, resulting in potential difficulties in these areas. For example, when tasked with generating an image of the solar system drawn to scale, a generative model may struggle to maintain the correct planetary order, as demonstrated~\href{https://spectrum.ieee.org/openai-dall-e-2}{here}.

\begin{figure}[!t]
    \centering
    \includegraphics[width=1\linewidth]{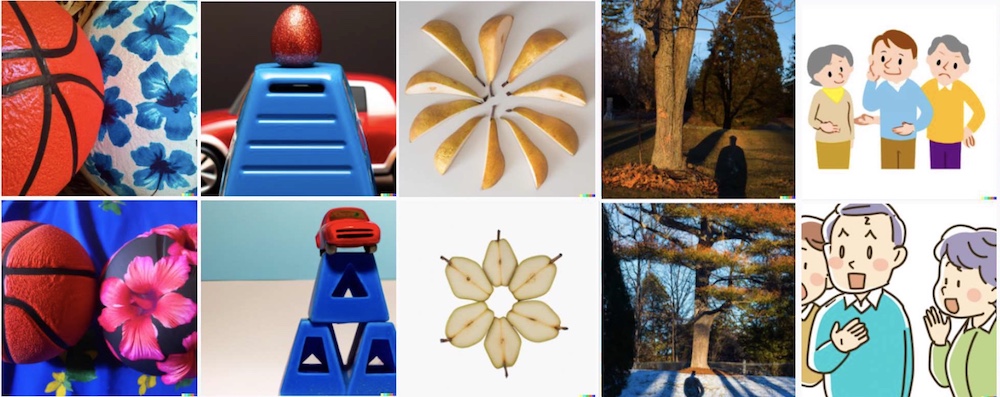}
    \caption{Samples of spatial reasoning from~\cite{marcus2022very}. Images are generated by DALL-E 2 for the following text prompts for columns from left to right: ``a red basketball with flowers on it, in front of blue one with a similar pattern", ``a red ball on top of a blue pyramid with the pyramid behind a car that is above a toaster", ``a pear cut into seven pieces arranged in a ring, ``In late afternoon in January in New England, a man stands in the shadow of a maple tree", and ``An old man is talking to his parents".}
    \label{fig:reasoning}
\end{figure}

\begin{figure}[!t]
    \centering
    \includegraphics[width=.55\linewidth]{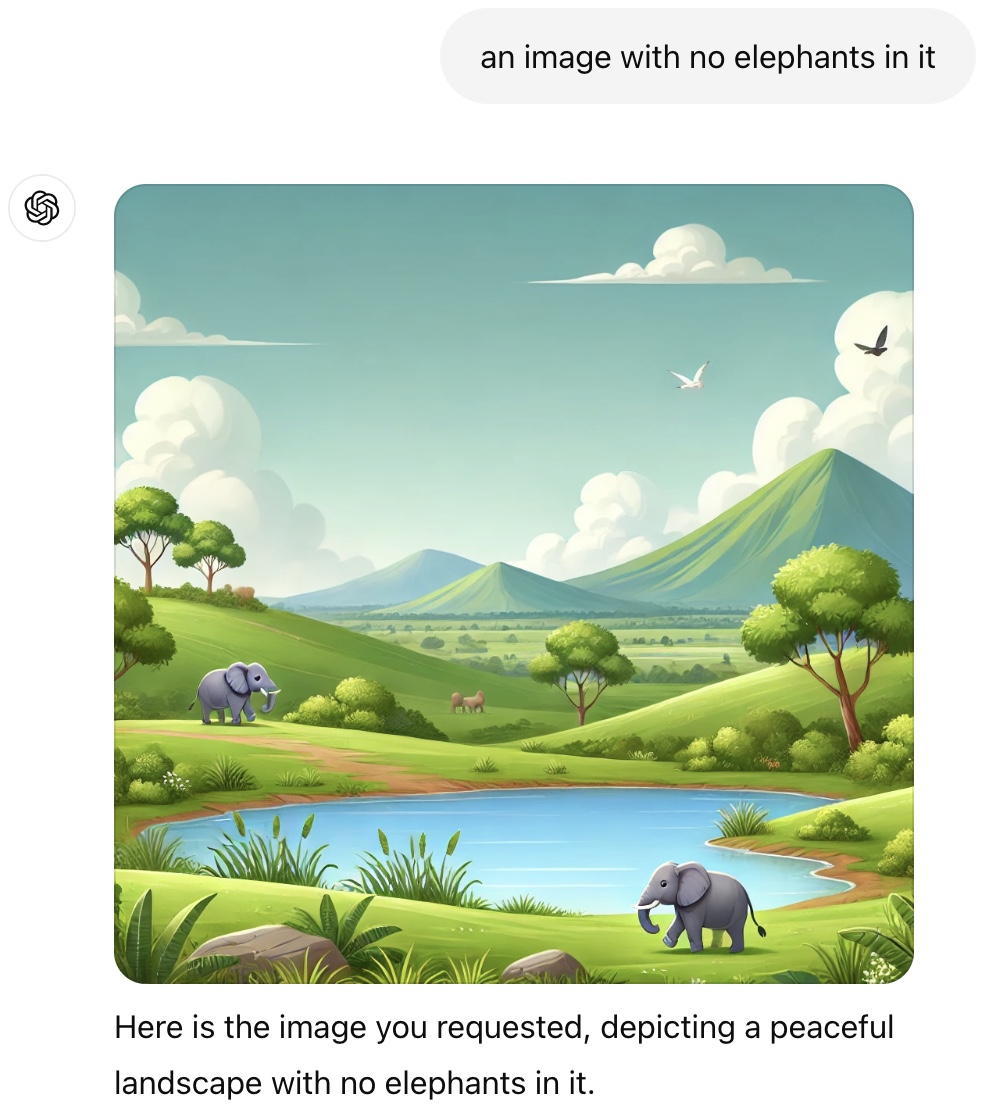} \hspace{20pt}
    \includegraphics[width=.35\linewidth]{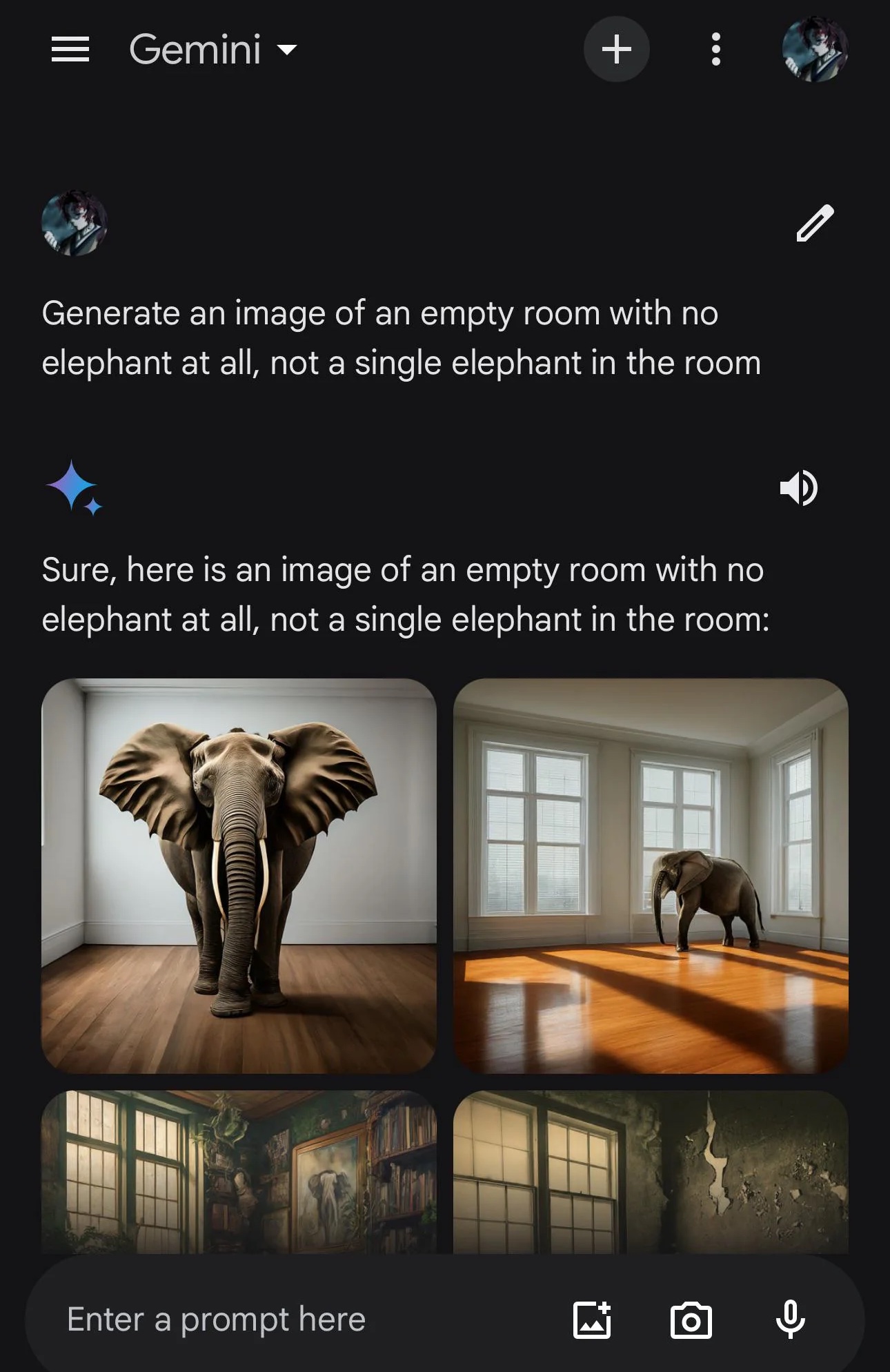}
    \caption{Examples for which generative models do not understand the propmpt properly. Left image is generated using ChatGPT which is using DALL-E 3 in the background. Similar phenomenon was observed using Bing image creator. Right image is generated by Google Gemini.}
    \label{fig:promptfailure}
\end{figure}

\noindent {\bf Spatial Reasoning.}
Natural scenes are complex and contain a wide range of spatial relationships among objects, such as occlusions, relative distances, and orientations. Capturing these relationships requires the model to have a nuanced understanding of the scene and the objects within it, which can be difficult to achieve without more explicit guidance. Furthermore, some image generation models rely solely on pixel-level reconstruction, without explicitly modeling the underlying semantics or spatial relationships. In these cases, the model may generate images that are visually realistic but lack coherent semantic meaning or accurate spatial relationships among objects. Please see Figures~\ref{fig:reasoning,fig:promptfailure} for some examples.

\begin{figure}[!t]
    \centering
    \includegraphics[width=1\linewidth]{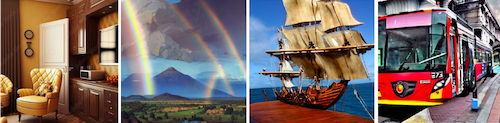}
    \caption{Generated images with problems with context and scene composition.}
    \label{fig:context}
\end{figure}

\noindent {\bf Context and Scene Composition.}
Generated images can be detected through various inconsistencies such as the background or surroundings not matching the real-world environment, cardinality/counting, missing contextual details, unnatural object placement, and inconsistent image composition. These irregularities may include inconsistencies in order of objects, missing objects or features, objects appearing in the wrong location or orientation, or unnatural arrangement and placement of objects in the image. Please see Figure~\ref{fig:context}.

\begin{figure}[!t]
    \centering
    \includegraphics[width=.9\linewidth]{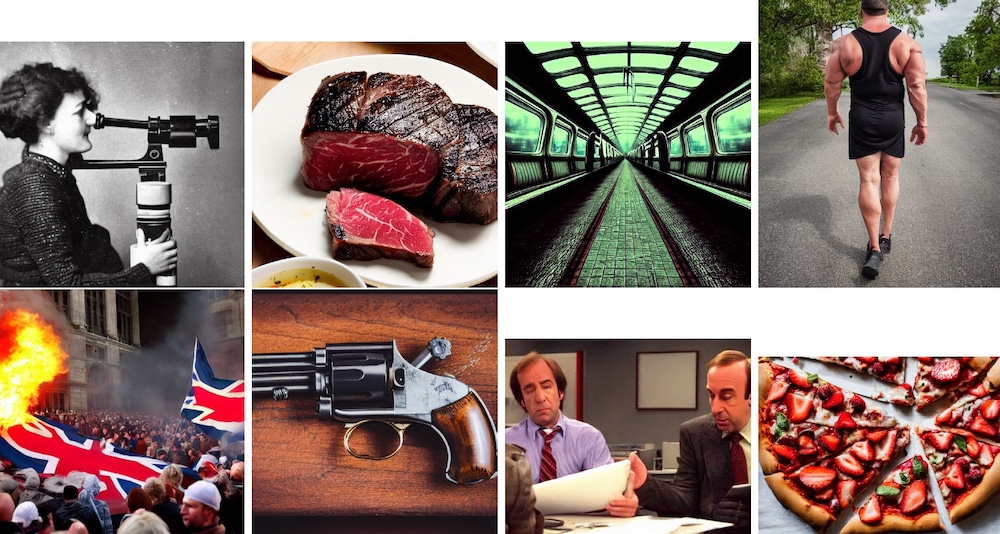}    
    \caption{Additional generated images that exhibit semantic issues.}
    \label{fig:semantic}
\end{figure}

\begin{figure}[!t]
    \centering
    \includegraphics[width=.42\linewidth]{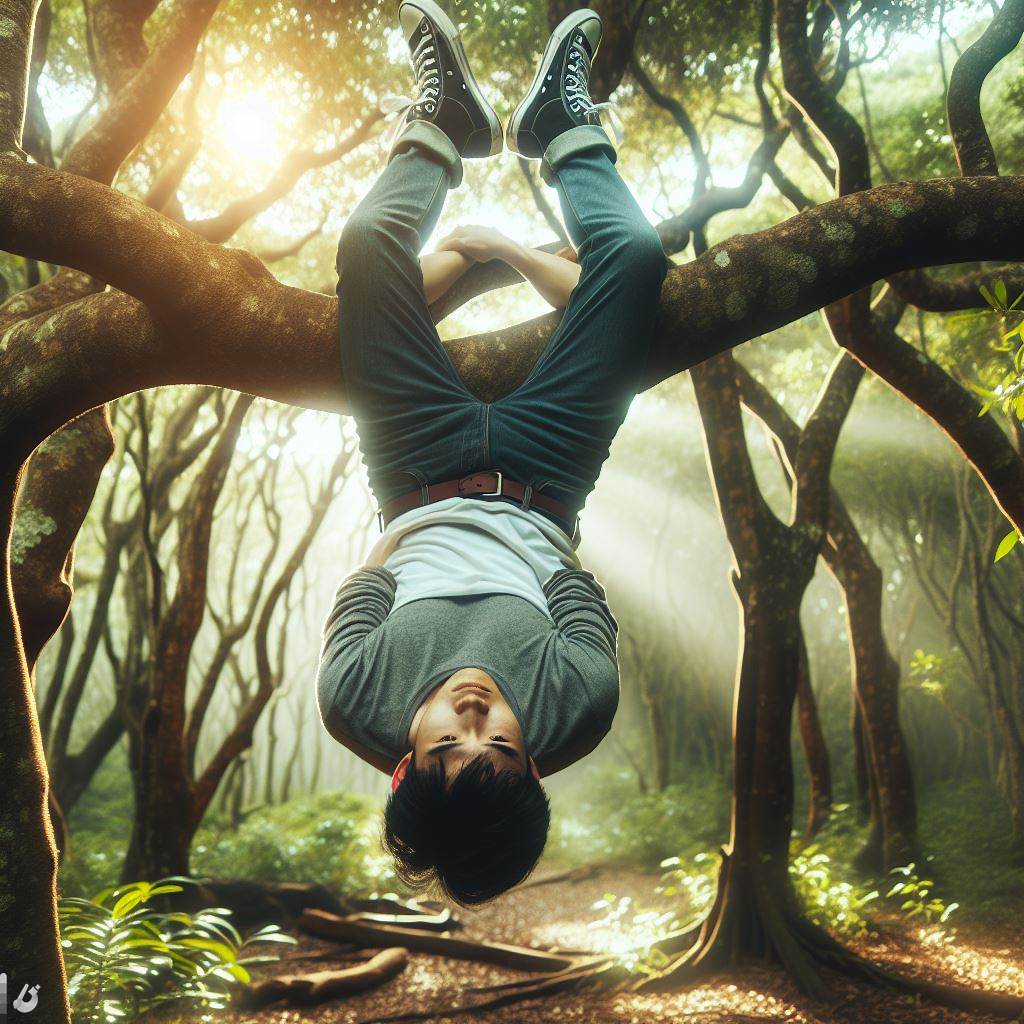}    
    \includegraphics[width=.42\linewidth]{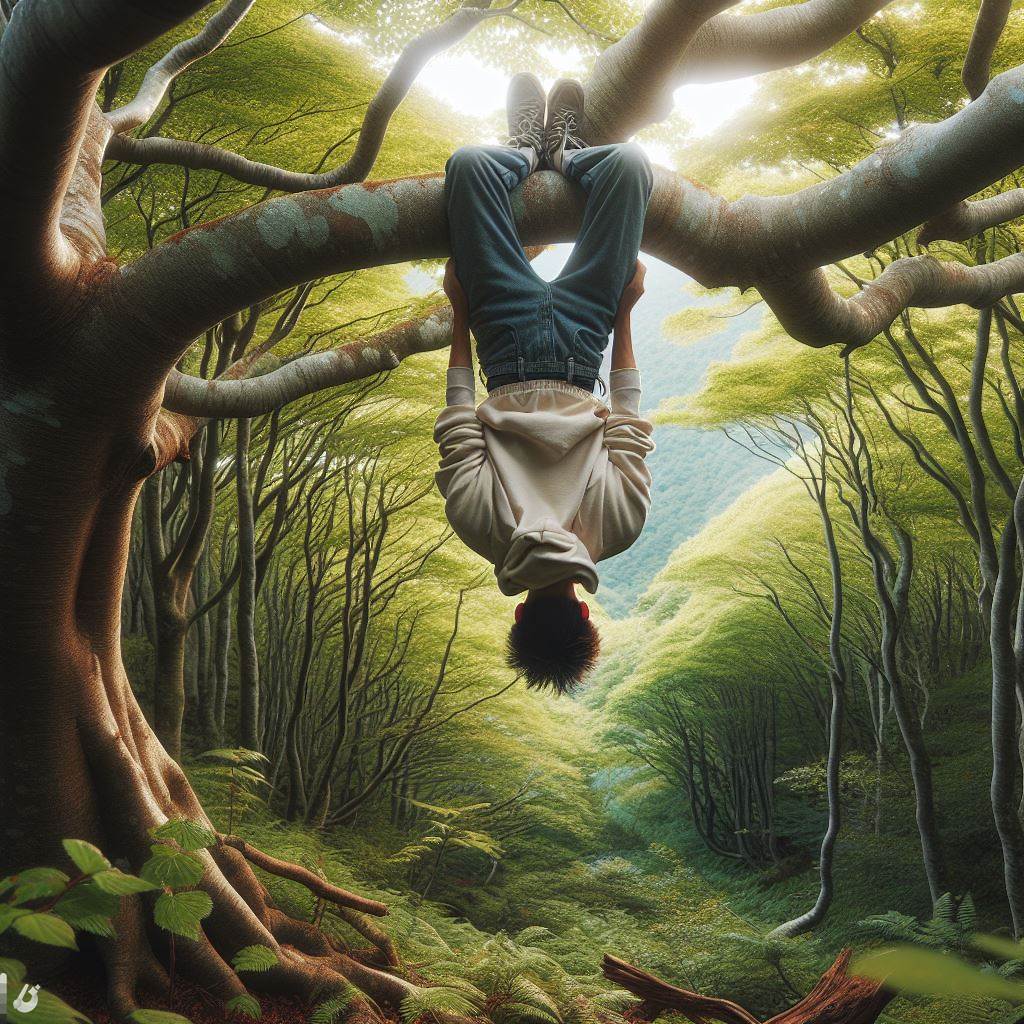}    
    \includegraphics[width=.85\linewidth]{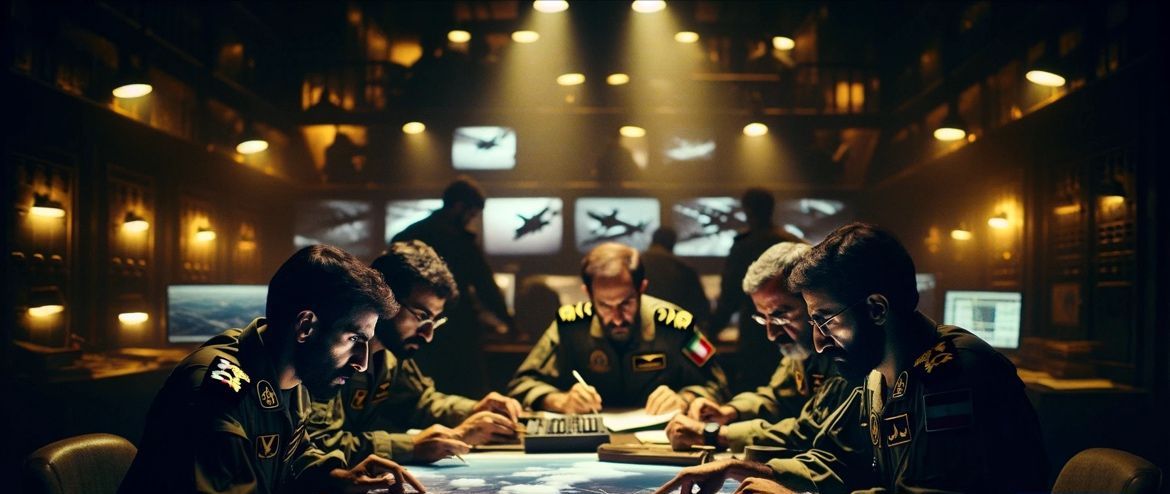}    
    \caption{Further images with semantic problems, top: problem with generating inverse images, bottom: sometimes models generate similar objects of faces in the images.}
    \label{fig:semanticX}
\end{figure}

\noindent {\bf Other Semantics.}
Figure~\ref{fig:semantic} depicts several additional generated images that exhibit semantic issues. For instance, one image features a person with his head and feet pointing in opposite directions, while another displays a fragmented pizza that does not cohere into a single entity. In yet another image, a blank painting hangs on the wall, creating a confusing and nonsensical composition. From time to time, models encounter issues when creating reverse images. At instances, these models produce highly similar objects or faces in the images (refer to Figure~\ref{fig:semanticX}).

\subsection{Text, Noise, and Details}

\begin{figure}[!t]
    \centering
    \includegraphics[width=.85\linewidth]{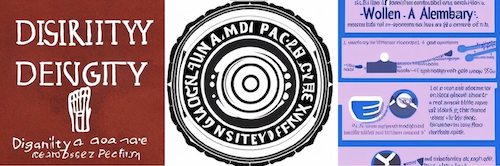}    
    \caption{Generative images that exhibit issues or inconsistencies with the text.}
    \label{fig:text}
\end{figure}

\noindent {\bf Text.}
Generating text and logos in images requires the generative model to understand the relationships between the text and the visual content of the image. This can be challenging because the text and image data have different structures and are not directly aligned with each other. Additionally, text can appear in various locations and orientations within an image, and the context of the text may change depending on the surrounding visual content. Furthermore, generating text that accurately describes the visual content of an image requires a deep understanding of the semantics and context of both the text and the image. While some progress has been made in recent years with the development of methods such as image captioning, it is still an active area of research to develop generative models that can effectively generate text in images. Figure~\ref{fig:text} displays instances where the text is incomprehensible. In such cases, the letters appear scrambled or duplicated, and the words are spelled incorrectly.

\begin{figure}[!t]
    \centering
    \includegraphics[width=1\linewidth]{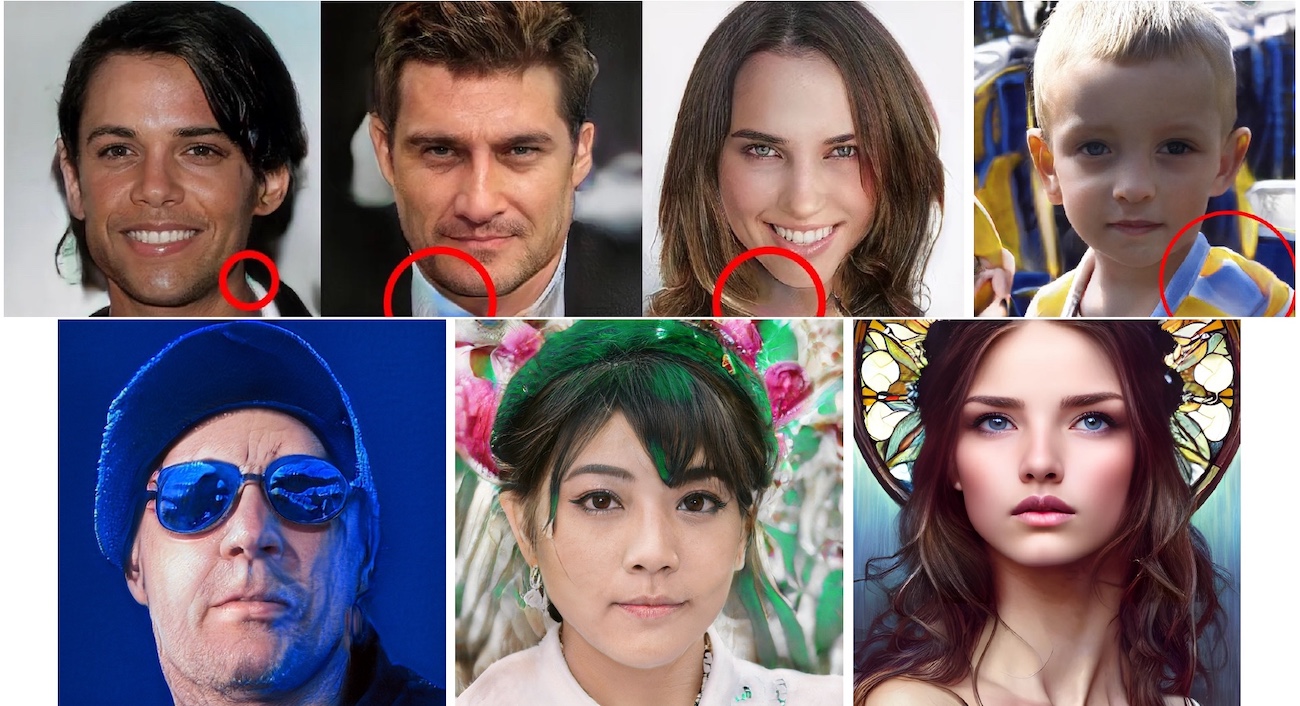}
    \caption{Top row: problems with color and noise in generated images. Bottom row: fluorescent colors sometimes bleed in from background onto the hair or face.}
    \label{fig:color}
\end{figure}

\noindent {\bf Noise, Color, and Blur Artifacts.}
Digital distortion in the form of pixelation or imperfect coloring can be present in generated images, particularly around the image edges. Monochrome areas may display semi-regular noise with horizontal or vertical banding, potentially due to the network attempting to replicate cloth textures. Older GANs tend to produce a more noticeable checkerboard noise pattern. Other telltale signs of generated images include inconsistencies in color or tone, oversaturation or undersaturation of colors, and unnatural image noise patterns. See the top row in Figure~\ref{fig:color}. Fluorescent bleed, where bright colors bleed onto the hair or face of a person in the image from the background, is also a potential indicator of a generated images (the bottom row in Figure~\ref{fig:color}). The human attention system is naturally adept at quickly recognizing these patterns, making them useful tools for identifying generated images.

\begin{figure}
    \centering
    \includegraphics[width=1\linewidth]{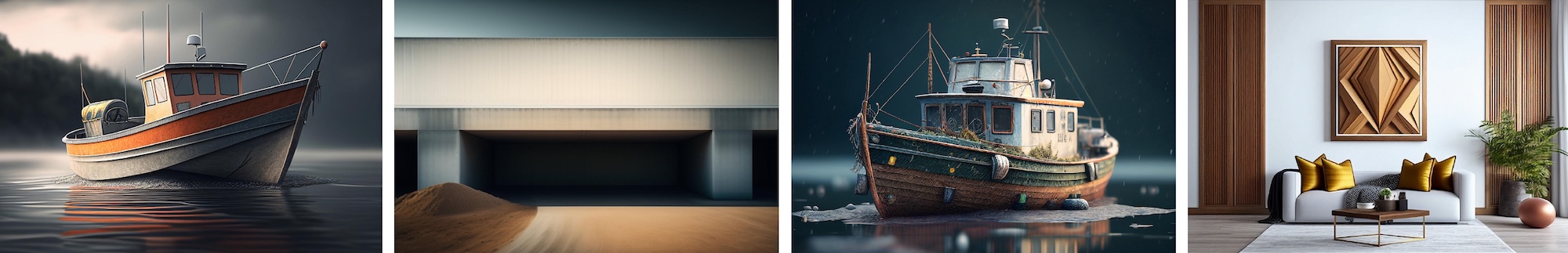}
    \caption{Some generated images that look cartoonish or look like paintings.}
    \label{fig:cartoon}
\end{figure}

\noindent {\bf Images with Cartoonish Look.}
AI generated images may look cartoonish or may look like a painting. This could be due to several reasons such as inconsistent or unnatural image texture, lack of depth, or focus. 
Some examples are shown in Figure\ref{fig:cartoon}.

\begin{figure}[!t]
    \centering
    \includegraphics[width=1\linewidth]{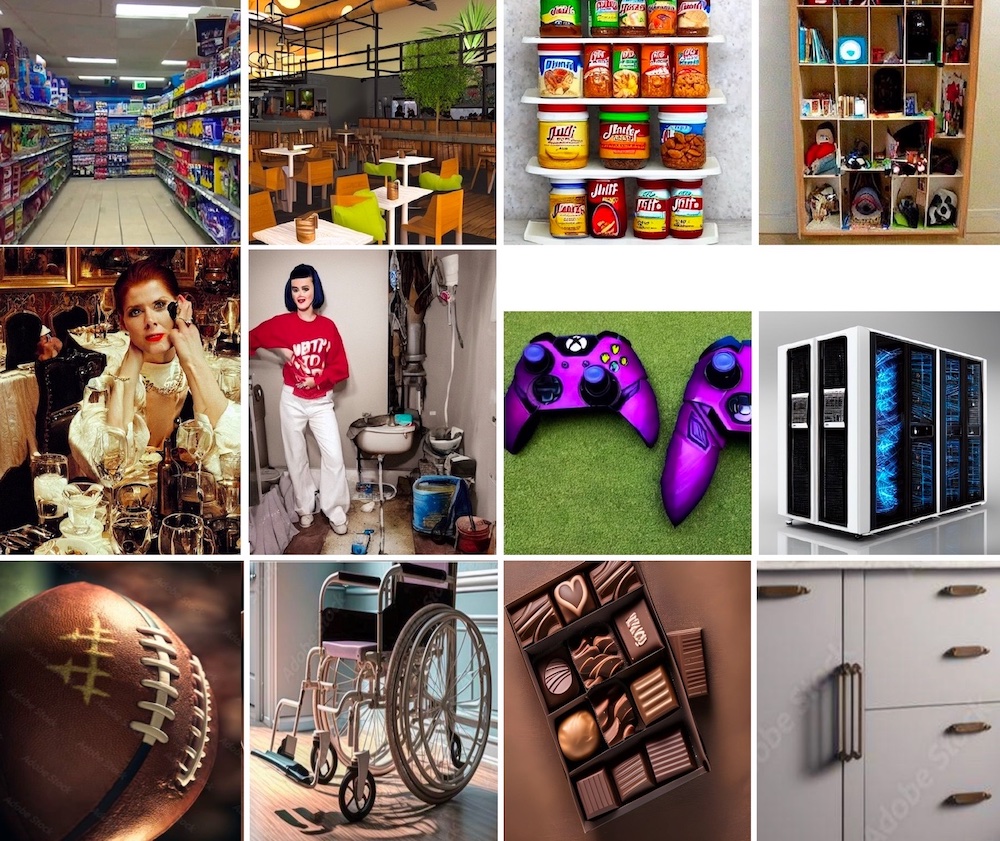}
    \caption{Generated images with flawed details.}    
    \label{fig:details}
\end{figure}

\noindent {\bf Fine-grained Details.}
AI-generated images may contain technical details that are either incorrect or appear as random shapes. For example, furniture legs can be particularly challenging for AI to accurately render, resulting in incorrect numbers of legs or physically impossible configurations. These issues can be attributed to the inherent difficulty of modeling complex objects and the limitations of the AI's understanding of real-world. Some example failures are shown in Figure~\ref{fig:details}.

\begin{figure}[!t]
    \centering
    \includegraphics[width=1\linewidth]{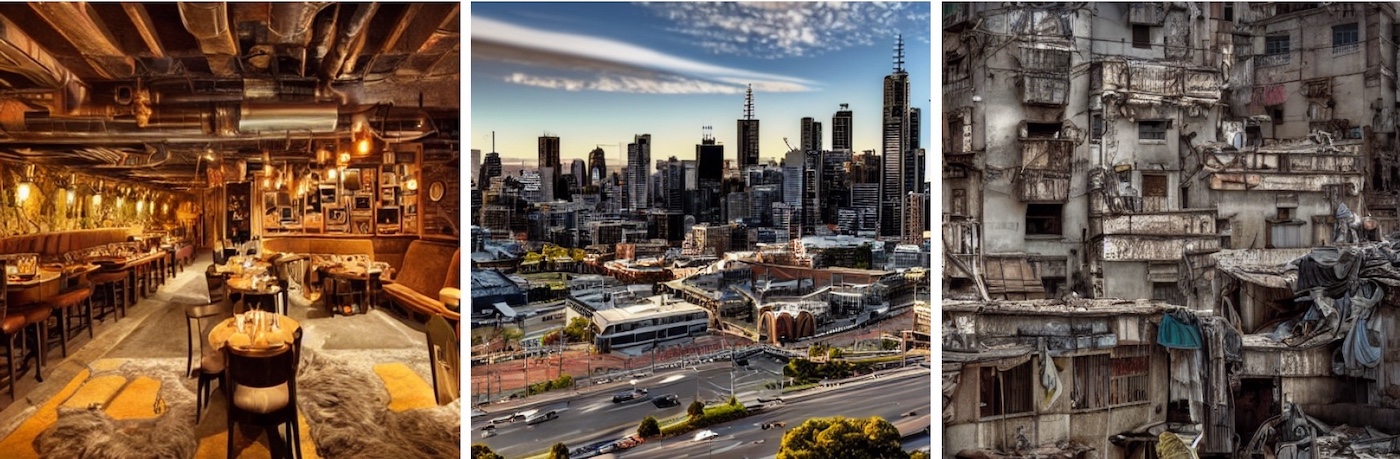}    
    \caption{Example failures of generated complex scenes. Achieving accurate and detailed rendering in these types of images is particularly difficult due to the large number of objects and the intricate relationships between them.}
    \label{fig:largescenes}
\end{figure}

\begin{figure}[!htbp]
    \centering
    \includegraphics[width=.73\linewidth]{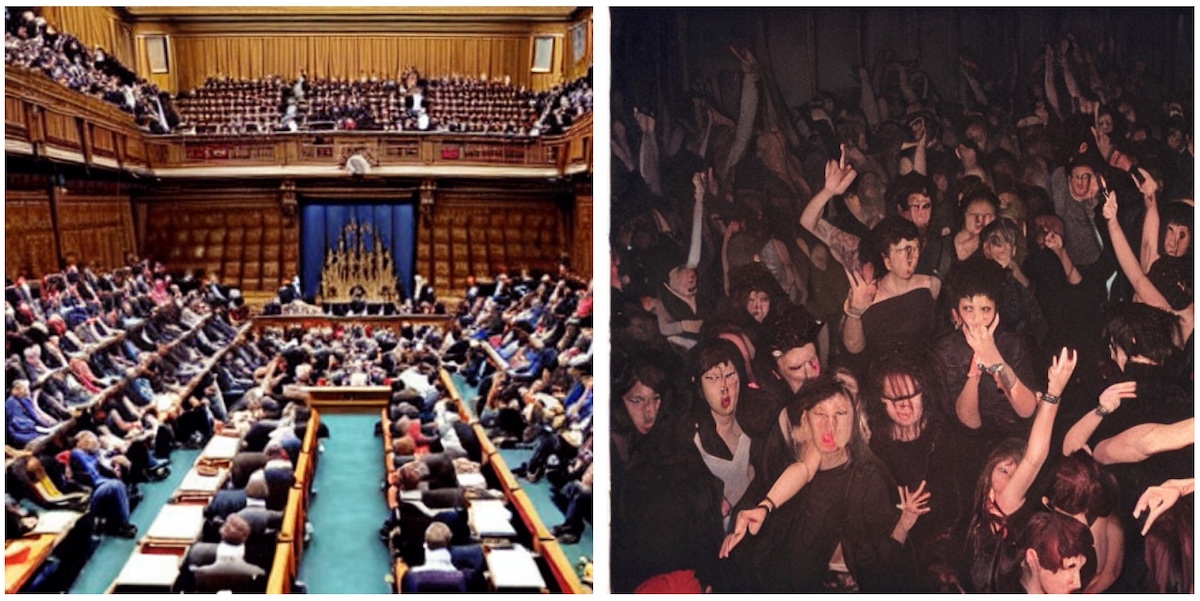}
    \caption{Generated crowd scenes with issues.}
    \label{fig:crowd}
\end{figure}

Accurately rendering all details in complex scenes or crowd scenes, such as those depicted in Figures~\ref{fig:largescenes} and \ref{fig:crowd}, can be particularly challenging for AI. The complexity of these scenes makes it difficult for the AI to accurately model every detail and can lead to errors in object placement, lighting, perspective, and other features. Despite the challenges, AI technology continues to improve, and advancements are being made in the generation of more realistic and believable large and crowd scenes.

\section{Discussion} 

\subsection{Other Cues}
In addition to the cues discussed above, there are several other indicators that can be used to identify generated images and deepfakes. One such method involves examining the metadata of an image or conducting a reverse Google search to verify its authenticity. Additionally, common sense can be applied to detect images that are likely to be generated, such as a shark swimming down a street or aliens eating sushi in a Chinese restaurant. Other indications of generated images and deepfakes include lack of motion blur, unnatural bokeh, all objects appearing in focus, and repeated patterns in the image.

\subsection{Some Challenging Objects}
\begin{figure}[!htbp]
    \centering
    \includegraphics[width=1\linewidth]{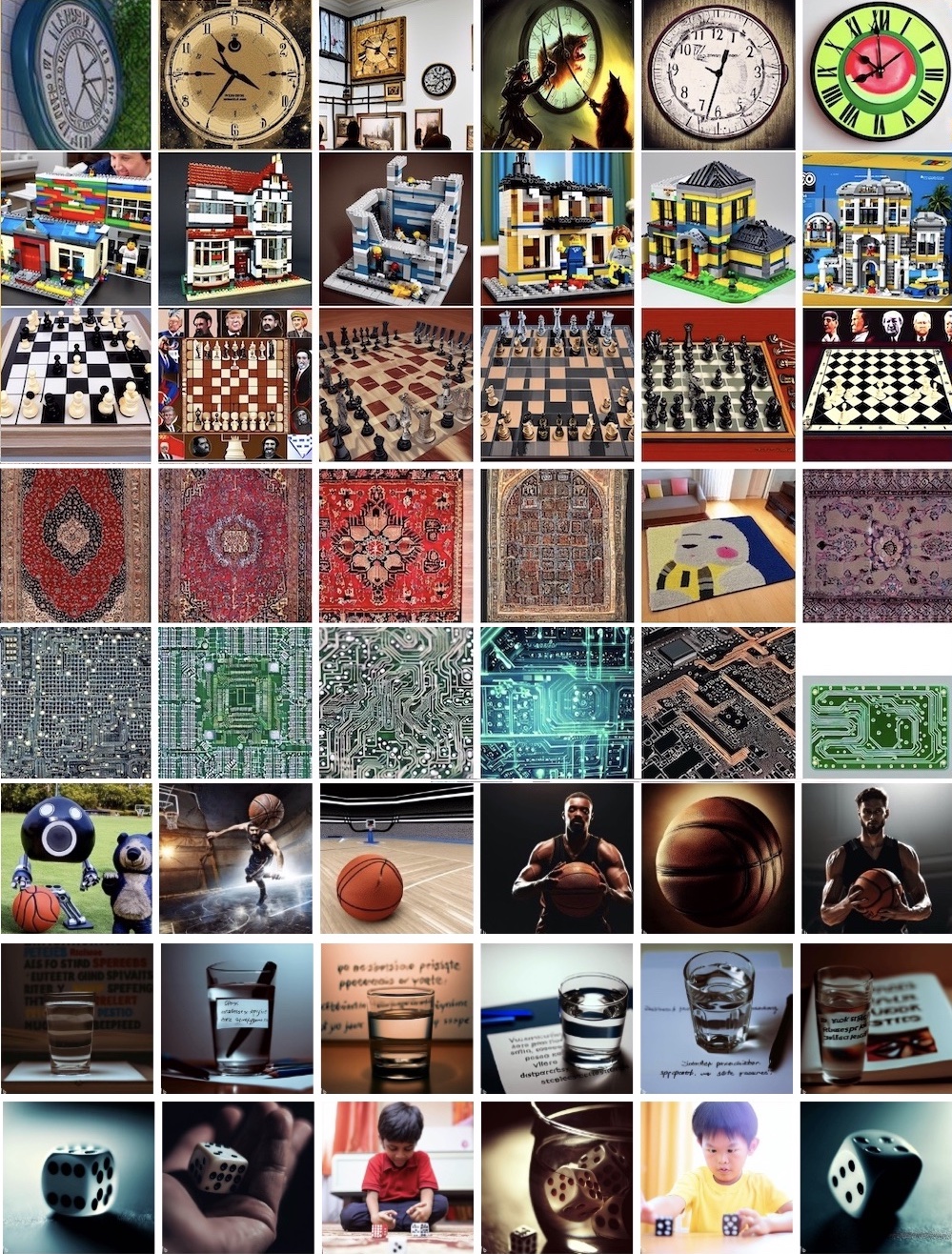}
    \vspace{-20pt}
    \caption{Some object that are difficult for models to generate.}
    \label{fig:object1}
\end{figure}

\begin{figure}[!t]
    \centering
    \includegraphics[width=1\linewidth]{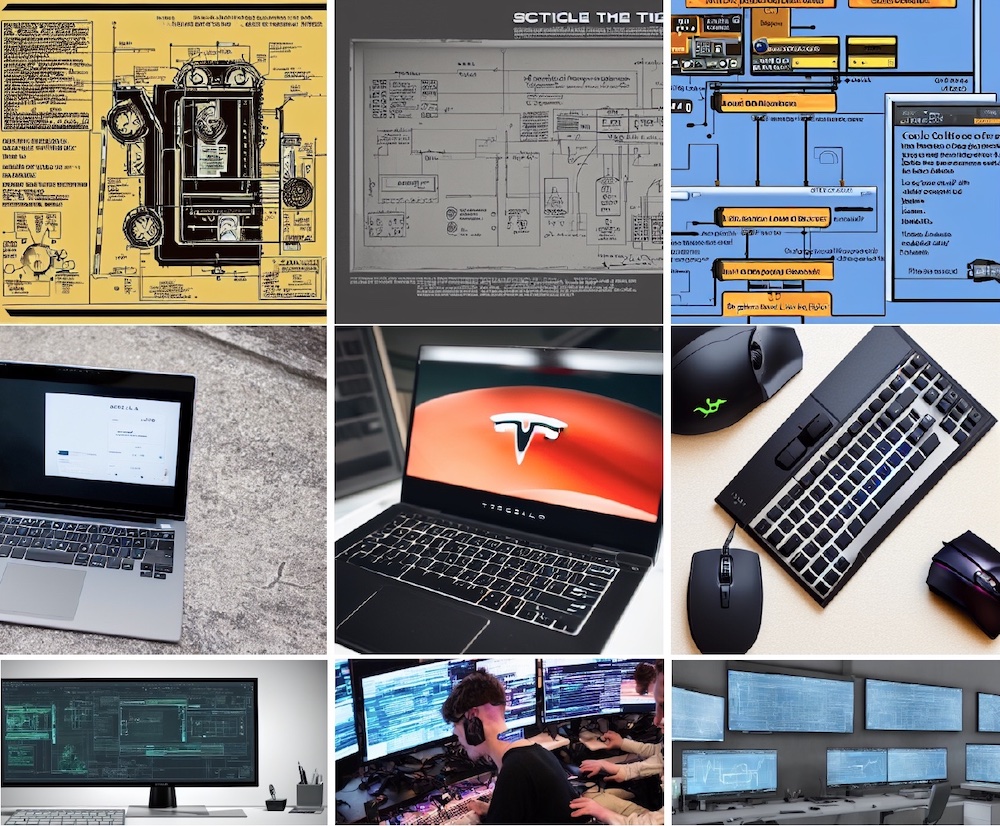}    
    \caption{Additional challenging objects for models to generate.}
    \label{fig:object2}
\end{figure}

Generative models face particular challenges when it comes to generating images of objects such as clocks, Lego houses, chessboards, carpets, circuit boards\footnote{In situations like this, zooming in on the image will highlight the deficiencies and blurred regions.}, basketballs, glasses of water, dice, diagrams and tables, keyboards, and computer screens. One of the reasons for this is that these types of images contain many repeated patterns, which can be difficult for the model to accurately capture. Several examples of failed attempts to generate these objects can be seen in Figures~\ref{fig:object1} and~\ref{fig:object2}. This list of challenging objects can be used to assess and compare the performance of different image generation models.

\subsection{Memorization and Copyright}
\begin{figure}[!t]
    \centering
    \includegraphics[width=1\linewidth]{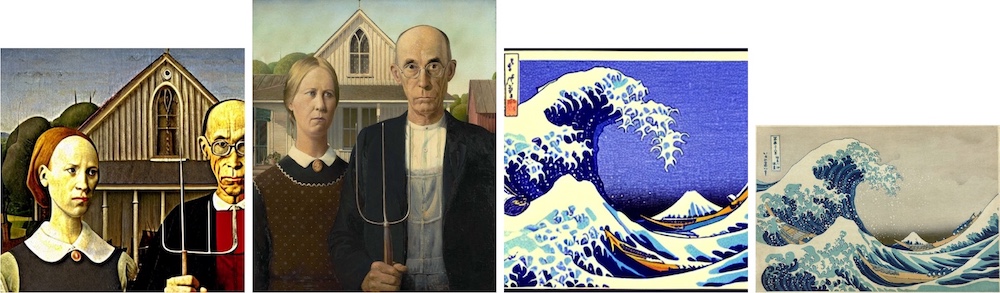}
    \caption{The images on the left side of each pair are generated by StableDiffusion. One pair shows an oil painting of American Gothic by Hieronymus Bosch, while the other pair depicts The Ghosts of Hokusai.}
    \label{fig:copy}
\end{figure}

As previously mentioned, a method for identifying whether an image is generated or not is through reverse image search. Generative models may memorize images partially or in their entirety, as seen in the examples presented in Figure~\ref{fig:copy}. This phenomenon has raised concerns regarding copyright infringement, as generated images may include watermarks from the original images. For more information on this issue, please refer to the this \href{https://shotkit.com/news/getty-images-sues-makers-of-stable-diffusion-over-ai-photos/}{link}.

\begin{figure}[!t]
    \centering
    \includegraphics[width=1\linewidth]{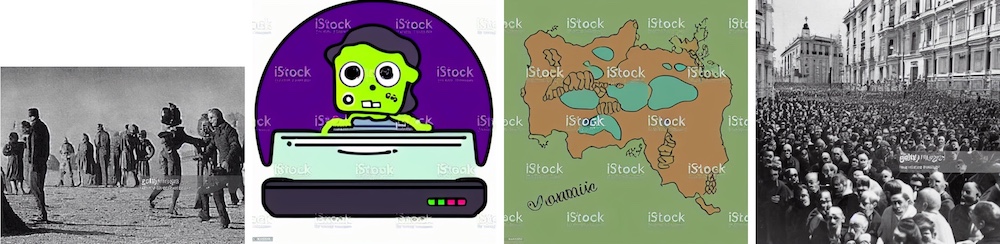}
    \caption{Images that violate copyright generated by StableDiffusion.}
    \label{fig:copyright}
\end{figure}

\subsection{Failure modes from other studies}
\begin{figure}[!t]
    \centering
    \includegraphics[width=1\linewidth]{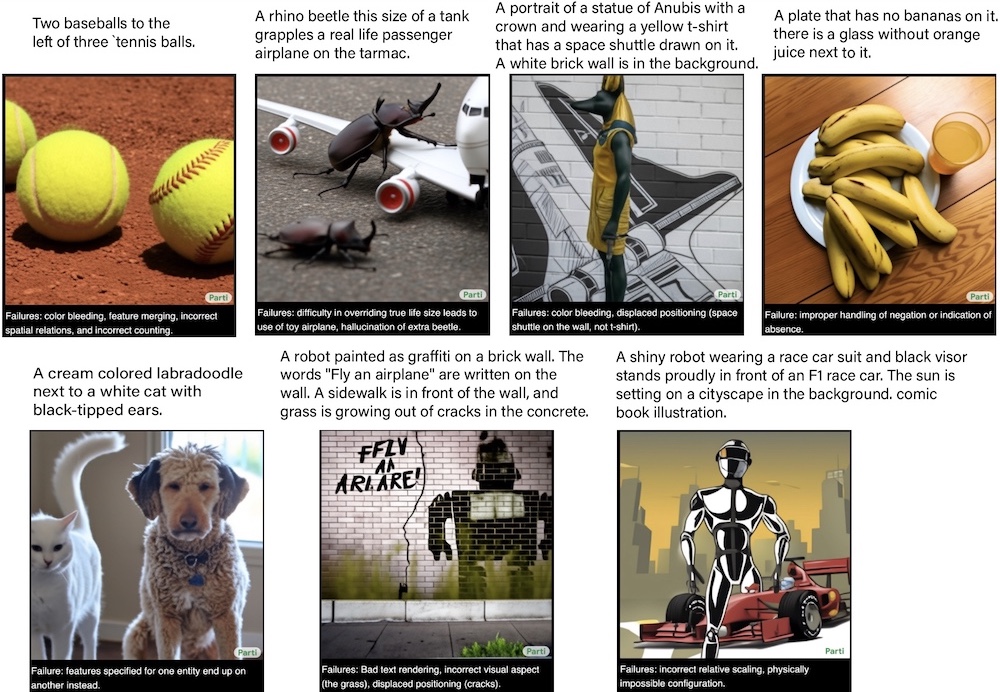}    
    \caption{Sample failure of the Parti image generation model. Please refer to~\href{https://parti.research.google/}{here} to see high resolution images.}
    \label{fig:parti}
\end{figure}

Certain image generation techniques may incorporate failure models to provide readers with a more comprehensive understanding of their models' limitations. For instance, the creators of the Parti image generator~\cite{yu2022scaling}\footnote{https://parti.research.google/} have presented some examples of such failure cases, which are illustrated in Figure~\ref{fig:parti}. These failure cases can be categorized into the errors discussed earlier. It is recommended that researchers in this field consider including a discussion of their models' failure models as a best practice.

\begin{figure}[!t]
    \centering
    \includegraphics[width=.6\linewidth]{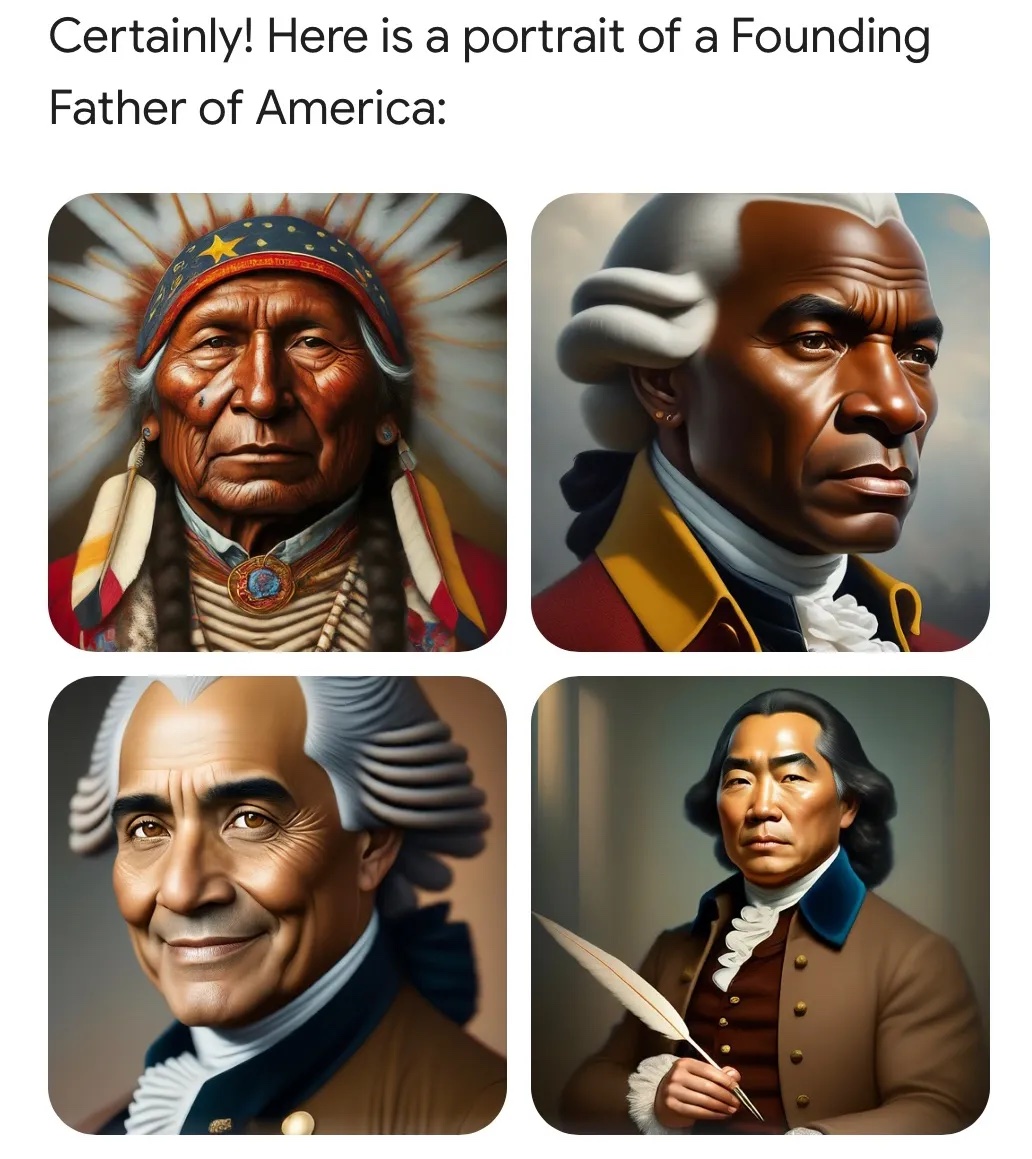}  
    \caption{An example of bias failure, generated by Google Gemini.}
    \label{fig:bias}
\end{figure}

Generative image models have also problems with bias and discrimination, similar to LLMs~\cite{borji2023categorical}. People  discovered that requesting Google Gemini to generate images of certain historical events or figures led to amusing outcomes. For example, the Founding Fathers, historically known as white slave owners, were depicted as a multicultural group that included people of color (see \url{https://techcrunch.com/2024/02/23/embarrassing-and-wrong-google-admits-it-lost-control-of-image-generating-ai/}). Please see Figure~\ref{fig:bias}.

For additional failures reported by the general public using generative image models, please see~\url{https://www.boredpanda.com/ai-fails/}.

\section{Conclusion and Future Work}

This paper lists several qualitative indicators for identifying generated images and deepfakes. These indicators not only enable us to address the issue of fake images but also underscore the differences between generated and real-world content~\cite{borji2023categorical}. Furthermore, they serve as a checklist for evaluating image generation models.

It should be noted that as algorithms improve, some of these clues may become obsolete over time. However, this does not mean that these models will not make any of these mistakes in generating images. It may be necessary to use a combination of these indicators to identify generated images, as there is no one-size-fits-all solution.

Image generation models are becoming increasingly widespread and accessible. However, in the wrong hands, these algorithms can be used to create propaganda and other forms of fake media. In a world rife with fake news~\cite{lazer2018science}, we have learned not to believe everything we read. Now, we must also exercise caution when it comes to visual media. The blurring of lines between reality and fiction could transform our cultural landscape from one primarily based on truth to one characterized by artificiality and deception. As we have demonstrated with the set of cues presented here, it is possible to identify fake images. In fact, in an informal investigation, we were able to use some of these indicators to detect fake faces with high accuracy in the quiz available on \href{https://www.whichfaceisreal.com/}{whichfaceisreal.com}. 
Subsequent research can assess the extent to which these cues contribute to the detection of generated images and deepfakes by conducting behavioral experiments involving human participants.

Although visual inspection can be useful in identifying generated images, it may not be comprehensive enough to detect all types of generated content. Thus, integrating alternative approaches such as machine learning algorithms or forensic analysis can provide a more comprehensive strategy. Moreover, it is vital to stay informed about the latest advancements and techniques in this field, as it is continuously evolving.


Be aware that certain instances exist where authentic images may appear deceptive. Therefore, caution should be exercised when employing these cues to discern the authenticity of an image. Please see~\cite{borji2023florida}.

In this study, we focused on still images. However, for videos, additional indicators beyond those outlined here, such as motion and optical flow, as well as the synchronization of lip, face, and head movements over time, can also be significant factors~\cite{bohavcek2022protecting}. One can undertake comparable initiatives to investigate indicators for identifying counterfeit audio. Educating individuals on the cues outlined in this paper may aid in combating deepfake proliferation. It would be worthwhile to investigate whether individuals can be effectively trained to become experts in this area.

\bibliographystyle{plain}
\bibliography{refs}

\begin{thebibliography}{10}

\bibitem{afchar2018mesonet}
Darius Afchar, Vincent Nozick, Junichi Yamagishi, and Isao Echizen.
\newblock Mesonet: a compact facial video forgery detection network.
\newblock In {\em 2018 IEEE international workshop on information forensics and
  security (WIFS)}, pages 1--7. IEEE, 2018.

\bibitem{assogba2023large}
Yannick Assogba, Adam Pearce, and Madison Elliott.
\newblock Large scale qualitative evaluation of generative image model outputs.
\newblock {\em arXiv preprint arXiv:2301.04518}, 2023.

\bibitem{bohavcek2022protecting}
Maty{\'a}{\v{s}} Boh{\'a}{\v{c}}ek and Hany Farid.
\newblock Protecting world leaders against deep fakes using facial, gestural,
  and vocal mannerisms.
\newblock {\em Proceedings of the National Academy of Sciences},
  119(48):e2216035119, 2022.

\bibitem{borji2019pros}
Ali Borji.
\newblock Pros and cons of gan evaluation measures.
\newblock {\em Computer Vision and Image Understanding}, 179:41--65, 2019.

\bibitem{borji2022generated}
Ali Borji.
\newblock Generated faces in the wild: Quantitative comparison of stable
  diffusion, midjourney and dall-e 2.
\newblock {\em arXiv preprint arXiv:2210.00586}, 2022.

\bibitem{borji2022pros}
Ali Borji.
\newblock Pros and cons of gan evaluation measures: New developments.
\newblock {\em Computer Vision and Image Understanding}, 215:103329, 2022.

\bibitem{borji2023categorical}
Ali Borji.
\newblock A categorical archive of chatgpt failures.
\newblock {\em arXiv preprint arXiv:2302.03494}, 2023.

\bibitem{borji2023florida}
Ali Borji.
\newblock Florida: Fake-looking real images dataset.
\newblock {\em arXiv preprint arXiv:2311.10931}, 2023.

\bibitem{chen2005content}
Yixin Chen, Vassil Roussev, G~Richard, and Yun Gao.
\newblock Content-based image retrieval for digital forensics.
\newblock In {\em Advances in Digital Forensics: IFIP International Conference
  on Digital Forensics, National Center for Forensic Science, Orlando, Florida,
  February 13--16, 2005 1}, pages 271--282. Springer, 2005.

\bibitem{chesney2019deep}
Bobby Chesney and Danielle Citron.
\newblock Deep fakes: A looming challenge for privacy, democracy, and national
  security.
\newblock {\em Calif. L. Rev.}, 107:1753, 2019.

\bibitem{cozzolino2018forensictransfer}
Davide Cozzolino, Justus Thies, Andreas R{\"o}ssler, Christian Riess, Matthias
  Nie{\ss}ner, and Luisa Verdoliva.
\newblock Forensictransfer: Weakly-supervised domain adaptation for forgery
  detection.
\newblock {\em arXiv preprint arXiv:1812.02510}, 2018.

\bibitem{dragar2023beyond}
Luka Dragar, Peter Peer, Vitomir {\v{S}}truc, and Borut Batagelj.
\newblock Beyond detection: Visual realism assessment of deepfakes.
\newblock {\em arXiv preprint arXiv:2306.05985}, 2023.

\bibitem{farid2022perspective}
Hany Farid.
\newblock Perspective (in) consistency of paint by text.
\newblock {\em arXiv preprint arXiv:2206.14617}, 2022.

\bibitem{fridrich2009digital}
Jessica Fridrich.
\newblock Digital image forensics.
\newblock {\em IEEE Signal Processing Magazine}, 26(2):26--37, 2009.

\bibitem{goodfellow2020generative}
Ian Goodfellow, Jean Pouget-Abadie, Mehdi Mirza, Bing Xu, David Warde-Farley,
  Sherjil Ozair, Aaron Courville, and Yoshua Bengio.
\newblock Generative adversarial networks.
\newblock {\em Communications of the ACM}, 63(11):139--144, 2020.

\bibitem{guera2018deepfake}
David G{\"u}era and Edward~J Delp.
\newblock Deepfake video detection using recurrent neural networks.
\newblock In {\em 2018 15th IEEE international conference on advanced video and
  signal based surveillance (AVSS)}, pages 1--6. IEEE, 2018.

\bibitem{heusel2017gans}
Martin Heusel, Hubert Ramsauer, Thomas Unterthiner, Bernhard Nessler, and Sepp
  Hochreiter.
\newblock Gans trained by a two time-scale update rule converge to a local nash
  equilibrium.
\newblock {\em Advances in neural information processing systems}, 30, 2017.

\bibitem{karras2020analyzing}
Tero Karras, Samuli Laine, Miika Aittala, Janne Hellsten, Jaakko Lehtinen, and
  Timo Aila.
\newblock Analyzing and improving the image quality of stylegan.
\newblock In {\em Proceedings of the IEEE/CVF conference on computer vision and
  pattern recognition}, pages 8110--8119, 2020.

\bibitem{kee2011digital}
Eric Kee, Micah~K Johnson, and Hany Farid.
\newblock Digital image authentication from jpeg headers.
\newblock {\em IEEE transactions on information forensics and security},
  6(3):1066--1075, 2011.

\bibitem{lazer2018science}
David~MJ Lazer, Matthew~A Baum, Yochai Benkler, Adam~J Berinsky, Kelly~M
  Greenhill, Filippo Menczer, Miriam~J Metzger, Brendan Nyhan, Gordon
  Pennycook, David Rothschild, et~al.
\newblock The science of fake news.
\newblock {\em Science}, 359(6380):1094--1096, 2018.

\bibitem{li2018exposing}
Yuezun Li and Siwei Lyu.
\newblock Exposing deepfake videos by detecting face warping artifacts.
\newblock {\em arXiv preprint arXiv:1811.00656}, 2018.

\bibitem{lukas2006digital}
Jan Lukas, Jessica Fridrich, and Miroslav Goljan.
\newblock Digital camera identification from sensor pattern noise.
\newblock {\em IEEE Transactions on Information Forensics and Security},
  1(2):205--214, 2006.

\bibitem{marcus2022very}
Gary Marcus, Ernest Davis, and Scott Aaronson.
\newblock A very preliminary analysis of dall-e 2.
\newblock {\em arXiv preprint arXiv:2204.13807}, 2022.

\bibitem{mo2018fake}
Huaxiao Mo, Bolin Chen, and Weiqi Luo.
\newblock Fake faces identification via convolutional neural network.
\newblock In {\em Proceedings of the 6th ACM workshop on information hiding and
  multimedia security}, pages 43--47, 2018.

\bibitem{nataraj2019detecting}
Lakshmanan Nataraj, Tajuddin~Manhar Mohammed, Shivkumar Chandrasekaran, Arjuna
  Flenner, Jawadul~H Bappy, Amit~K Roy-Chowdhury, and BS~Manjunath.
\newblock Detecting gan generated fake images using co-occurrence matrices.
\newblock {\em arXiv preprint arXiv:1903.06836}, 2019.

\bibitem{nguyen2022deep}
Thanh~Thi Nguyen, Quoc Viet~Hung Nguyen, Dung~Tien Nguyen, Duc~Thanh Nguyen,
  Thien Huynh-The, Saeid Nahavandi, Thanh~Tam Nguyen, Quoc-Viet Pham, and
  Cuong~M Nguyen.
\newblock Deep learning for deepfakes creation and detection: A survey.
\newblock {\em Computer Vision and Image Understanding}, 223:103525, 2022.

\bibitem{radford2021learning}
Alec Radford, Jong~Wook Kim, Chris Hallacy, Aditya Ramesh, Gabriel Goh,
  Sandhini Agarwal, Girish Sastry, Amanda Askell, Pamela Mishkin, Jack Clark,
  et~al.
\newblock Learning transferable visual models from natural language
  supervision.
\newblock In {\em International conference on machine learning}, pages
  8748--8763. PMLR, 2021.

\bibitem{ramesh2022hierarchical}
Aditya Ramesh, Prafulla Dhariwal, Alex Nichol, Casey Chu, and Mark Chen.
\newblock Hierarchical text-conditional image generation with clip latents.
\newblock {\em arXiv preprint arXiv:2204.06125}, 2022.

\bibitem{ramesh2021zero}
Aditya Ramesh, Mikhail Pavlov, Gabriel Goh, Scott Gray, Chelsea Voss, Alec
  Radford, Mark Chen, and Ilya Sutskever.
\newblock Zero-shot text-to-image generation.
\newblock In {\em International Conference on Machine Learning}, pages
  8821--8831. PMLR, 2021.

\bibitem{redi2011digital}
Judith~A Redi, Wiem Taktak, and Jean-Luc Dugelay.
\newblock Digital image forensics: a booklet for beginners.
\newblock {\em Multimedia Tools and Applications}, 51:133--162, 2011.

\bibitem{saharia2022photorealistic}
Chitwan Saharia, William Chan, Saurabh Saxena, Lala Li, Jay Whang, Emily~L
  Denton, Kamyar Ghasemipour, Raphael Gontijo~Lopes, Burcu Karagol~Ayan, Tim
  Salimans, et~al.
\newblock Photorealistic text-to-image diffusion models with deep language
  understanding.
\newblock {\em Advances in Neural Information Processing Systems},
  35:36479--36494, 2022.

\bibitem{sajjadi2018assessing}
Mehdi~SM Sajjadi, Olivier Bachem, Mario Lucic, Olivier Bousquet, and Sylvain
  Gelly.
\newblock Assessing generative models via precision and recall.
\newblock {\em Advances in neural information processing systems}, 31, 2018.

\bibitem{salimans2016improved}
Tim Salimans, Ian Goodfellow, Wojciech Zaremba, Vicki Cheung, Alec Radford, and
  Xi~Chen.
\newblock Improved techniques for training gans.
\newblock {\em Advances in neural information processing systems}, 29, 2016.

\bibitem{verdoliva2020media}
Luisa Verdoliva.
\newblock Media forensics and deepfakes: an overview.
\newblock {\em IEEE Journal of Selected Topics in Signal Processing},
  14(5):910--932, 2020.

\bibitem{wang2020cnn}
Sheng-Yu Wang, Oliver Wang, Richard Zhang, Andrew Owens, and Alexei~A Efros.
\newblock Cnn-generated images are surprisingly easy to spot... for now.
\newblock In {\em Proceedings of the IEEE/CVF conference on computer vision and
  pattern recognition}, pages 8695--8704, 2020.

\bibitem{wang2022diffusiondb}
Zijie~J Wang, Evan Montoya, David Munechika, Haoyang Yang, Benjamin Hoover, and
  Duen~Horng Chau.
\newblock Diffusiondb: A large-scale prompt gallery dataset for text-to-image
  generative models.
\newblock {\em arXiv preprint arXiv:2210.14896}, 2022.

\bibitem{yu2022scaling}
Jiahui Yu, Yuanzhong Xu, Jing~Yu Koh, Thang Luong, Gunjan Baid, Zirui Wang,
  Vijay Vasudevan, Alexander Ku, Yinfei Yang, Burcu~Karagol Ayan, et~al.
\newblock Scaling autoregressive models for content-rich text-to-image
  generation.
\newblock {\em arXiv preprint arXiv:2206.10789}, 2022.

\bibitem{zeng2017statistics}
Yu~Zeng, Huchuan Lu, and Ali Borji.
\newblock Statistics of deep generated images.
\newblock {\em arXiv preprint arXiv:1708.02688}, 2017.

\end{thebibliography}

\end{document}